\pdfoutput=1

\documentclass[11pt]{article}

\usepackage[]{ACL2023}

\usepackage{times}
\usepackage{latexsym}
\usepackage{amsmath,amssymb,stmaryrd}

\usepackage[T1]{fontenc}

\usepackage[utf8]{inputenc}
\usepackage{paralist}

\usepackage{microtype}
\usepackage{booktabs,tabularx, colortbl}
\usepackage{graphicx}
\usepackage{amsmath}
\usepackage{multirow}
\usepackage{multibib}
\usepackage{makecell}
\usepackage{xspace}
\usepackage{colortbl}
\usepackage{ulem}

\usepackage{inconsolata}

\newcommand{\method}{\textsc{BigVideo}\xspace}
\newcommand{\howtwo}{\textsc{How2}\xspace}
\newcommand{\vatex}{\textsc{VaTeX}\xspace}
\newcommand{\testone}{\textsc{Ambiguous}\xspace}
\newcommand{\testtwo}{\textsc{Unambiguous}\xspace}

\title{\method: A Large-scale Video Subtitle Translation Dataset for Multimodal Machine Translation}

\author{Liyan Kang\textsuperscript{1,3}\footnotemark[1]~~~Luyang Huang\textsuperscript{2}\footnotemark[1]~~~Ningxin Peng\textsuperscript{2}~~~Peihao Zhu\textsuperscript{2}~~~Zewei Sun\textsuperscript{2}\\ {\bf Shanbo Cheng}\textsuperscript{2}~~~{\bf Mingxuan Wang}\textsuperscript{2}~~~{\bf Degen Huang}\textsuperscript{4}~~~{\bf Jinsong Su}\textsuperscript{1,3}\footnotemark[2]\\
  \textsuperscript{1}School of Informatics, Xiamen University ~~~
  \textsuperscript{2}Bytedance\\
  \textsuperscript{3}Key Laboratory of Digital Protection and Intelligent Processing of Intangible Cultural  \\  Heritage   of Fujian and Taiwan (Xiamen University), Ministry of Culture and Tourism, China\\
  \textsuperscript{4}Dalian University of Technology\\
  \texttt{\small{liyankang@stu.xmu.edu.cn}}~~~\texttt{\small{huangluyang@bytedance.com}}~~~\texttt{\small{jssu@xmu.edu.cn}}
  }
  
\begin{document}
\maketitle

\renewcommand{\thefootnote}{\fnsymbol{footnote}}
\footnotetext[1]{Equal contribution. Work was done while Liyan Kang was interning at ByteDance.}
\footnotetext[2]{Corresponding author.}
\renewcommand{\thefootnote}{\arabic{footnote}}

\begin{abstract}

We present a large-scale video subtitle translat-\\ion dataset, \method, to facilitate the study of multi-modality machine translation. 
Compared with the widely used \howtwo and \vatex datasets, 
\method is more than 10 times larger, consisting of 4.5 million sentence pairs and  9,981 hours of videos. 
We also introduce two deliberately designed test sets to verify the necessity of visual information:  \testone with the presence of ambiguous words, and \testtwo in which the text context is self-contained for translation. To better model the common semantics shared across texts and videos,  we introduce a contrastive learning method in the cross-modal encoder. Extensive experiments on the \method show that: 
\begin{inparaenum}[\it a)]
    \item Visual information consistently improves the NMT model in terms of BLEU, BLEURT, and COMET on both \testone and \testtwo test sets. 
    \item Visual information helps disambiguation, compared to the strong text baseline on terminology-targeted scores and human evaluation. Dataset and our implementations are available at 
    \href{https://github.com/DeepLearnXMU/BigVideo-VMT}{https://github.com/DeepLearnXMU/BigVideo-VMT}.
\end{inparaenum}

\end{abstract}

\section{Introduction}
Humans are able to integrate both language and visual context  to understand the world. 
From the perspective of NMT, it is also much needed to make use of such information to approach human-level translation abilities. 
To facilitate Multimodal Machine Translation (MMT) research, a number of datasets have been proposed including image-guided translation datasets~\cite{elliott-etal-2016-multi30k,gella-etal-2019-cross,wang-etal-2022-multisubs} and video-guided translation datasets~\cite{DBLP:journals/corr/abs-1811-00347,DBLP:conf/iccv/WangWCLWW19,li-etal-2022-visa}. 

However, the conclusion about the effects of visual  information is still unclear for MMT research~\cite{caglayan-etal-2019-probing}.  Previous work has suggested that visual information is only marginally beneficial for machine translation~\cite{li-etal-2021-vision,caglayan2021cross}, especially when the text context is not complete. The most possible reason is that existing datasets focus on captions describing images or videos, which are not large and diverse enough. 
The text inputs are often simple and sufficient for translation tasks~\cite{wu-etal-2021-good}. Take the widely used Multi30K as an example. Multi30K consists of only 30K image captions, while typical text translation systems are often trained with several million sentence pairs.

\begin{figure}[t]
    \centering
    \includegraphics[width=\columnwidth,trim=0 0 0 0, clip]{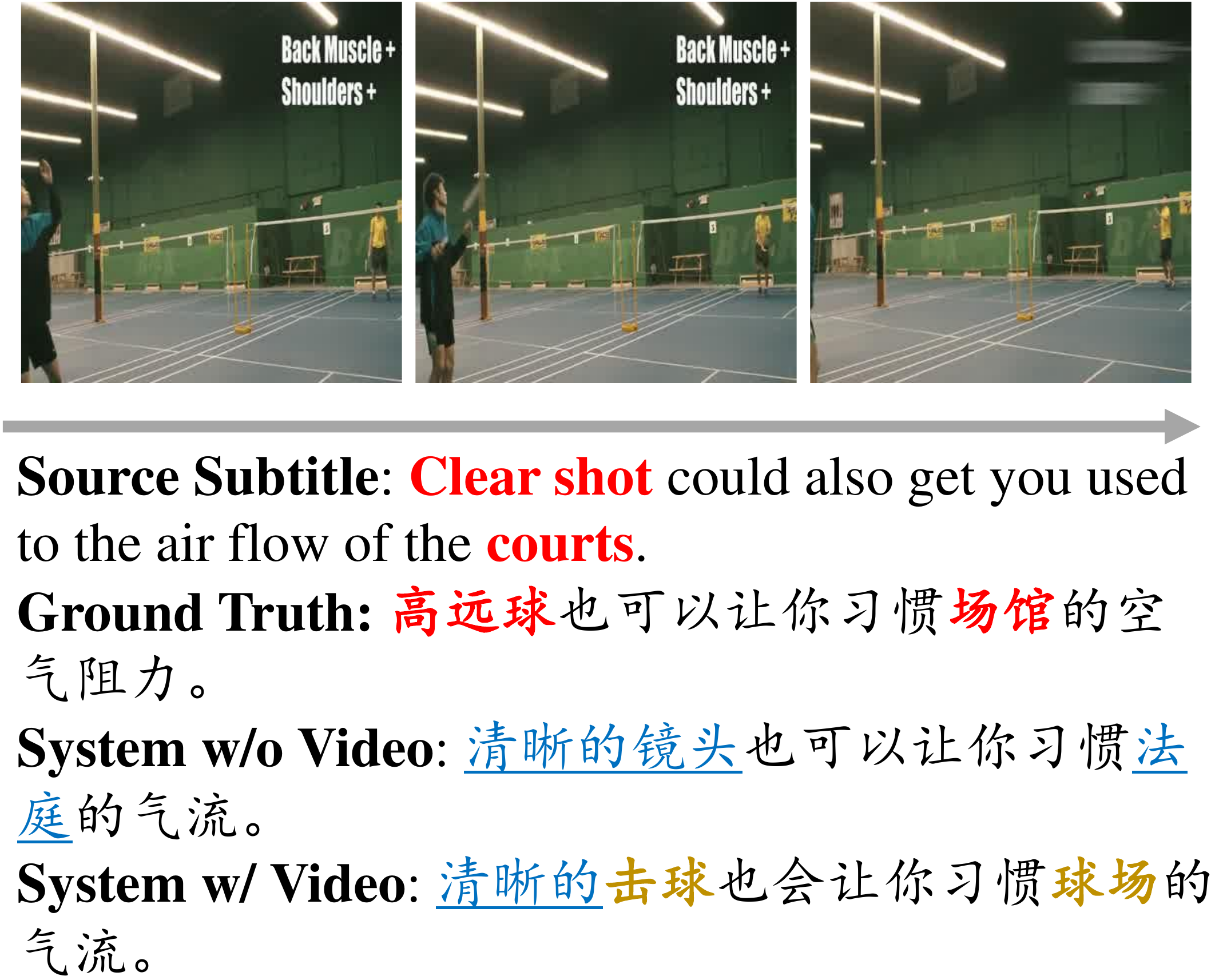}
    \caption{An example with semantic ambiguity in \method. The phrases with semantic ambiguity are highlighted in \textcolor{red}{\textbf{red}}. The wrong translations are in \textcolor[RGB]{0,112,192}{\underline{blue}} and the correct translations are in \textcolor[RGB]{191,144,0}{\textbf{yellow}}.
    }
    \label{fig:intro}
\end{figure}

We argue that studying the effects of visual contexts in machine translation requires a large-scale and diverse data set for training and a real-world and complex benchmark for testing. To this end, we propose \method, a large-scale video subtitle translation dataset. We collect human-written subtitles from two famous online video platforms, Xigua and YouTube. \method consists of 155 thousand videos and 4.5 million high-quality parallel sentences in English and Chinese. We highlight the key features of \method as
 follows:
\begin{inparaenum}[\it a)]
    \item The size of \method bypasses the largest available video machine translation dataset \howtwo and \vatex by one order of magnitude. 
    \item To investigate the need for visual information, two test sets are annotated by language experts, referred as \testone and \testtwo. In \testone, the source input is not sufficient enough  and requires videos to disambiguate for translation. The experts also labelled unambiguous words to help evaluate whether the improvement comes from visual contexts. In \testtwo, actions or visual scenes in the videos are mentioned in the subtitles but source sentences are self-contained for translation. 
\end{inparaenum}


To make the most of visual information for MMT, we propose a unified encoder-decoder framework for MMT. The model has a cross-modal encoder that takes both videos and texts as inputs. Motivated by recent work on cross-modal learning~\cite{li2020unicoder,qi2020imagebert,xia2021xgpt}, we also introduce a contrastive learning objective to further bridge the representation gap between the text and video and project them in a shared space. As such, the visual information can potentially contribute more to the translation model. 


We conduct extensive experiments on the proposed benchmark \method and report the results on BLEU~\cite{papineni-etal-2002-bleu}, BLEURT~\cite{sellam-etal-2020-bleurt}, COMET~\cite{rei-etal-2020-comet},  terminology-targeted metrics and human evaluation.  We also introduce the large scale WMT19 training data, which contains 20.4M parallel sentences to build the strong baseline model. 
 The experiments show that visual contexts consistently improve the performance of both the \testone and \testtwo test set over the strong text-only model. 
The finding is slightly different with previous studies and address the importance of a large scale and high-quality video translation data.
Further, the contrastive learning method can further boost the translation performance over other visual-guided models, which shows the benefits of closing the representation gap of texts and videos.

\section{Related Work}

\textbf{Video-guided Machine Translation.} The \vatex dataset has been introduced for video-guided machine translation task~\cite{DBLP:conf/iccv/WangWCLWW19}. It contains 129K bilingual captions paired with video clips. However, as pointed out by~\newcite{DBLP:journals/jip/YangHKO22}, captions in \vatex have sufficient information for translation, and models trained on \vatex tend to ignore video information. Beyond captions, \citet{DBLP:journals/corr/abs-1811-00347} considers video subtitles to construct the \howtwo dataset. \howtwo collects instructional videos from YouTube and obtains 186K bilingual subtitle sentences. To construct a challenging VMT dataset, \citet{li-etal-2022-visa} collect 39K ambiguous subtitles from movies or TV episodes to build \textsc{VISA}. However, both \howtwo and \textsc{VISA} are limited on scale and diversity, given the training needs of large models. In contrast, we release a larger video subtitle translation dataset, with millions of bilingual ambiguous subtitles, covering all categories on YouTube and Xigua platforms.

To leverage video inputs in machine translation models,~\newcite{DBLP:journals/corr/abs-2006-12799} use pretrained models such as ResNet~\cite{DBLP:conf/cvpr/HeZRS16}, Faster-RCNN~\cite{DBLP:conf/nips/RenHGS15} and I3D~\cite{DBLP:conf/cvpr/CarreiraZ17}. An additional attention module is designed in the RNN decoder to fuse visual features. To better learn temporal information in videos,~\cite{gu-etal-2021-video} propose a hierarchical attention network to model video-level features. Different from previous work, we use a unified encoder to learn both video and text features. Specifically, a contrastive learning objective is adopted to learn cross-modal interaction.

\noindent \textbf{Image-guided Machine Translation.} Images as additional inputs have long been used for machine translation~\cite{hitschler-etal-2016-multimodal}. For neural models, several attempts have been focused on enhancing the sequence-to-sequence model with strong image features~\cite{elliott-kadar-2017-imagination,yao-wan-2020-multimodal,DBLP:conf/mm/LinMSYYGZL20,DBLP:conf/acl/YinMSZYZL20,DBLP:journals/isci/SuCJZLGWL21,li-etal-2022-vision,DBLP:conf/mm/LinMSYYGZL20,DBLP:journals/corr/abs-2212-10313,ocracl2023}. However,~\newcite{li-etal-2021-vision} and~\newcite{wu-etal-2021-good} point out that images in Multi30K provide little information for translation. In this work, we focus on videos as additional visual inputs for subtitle translation. Videos illustrate objects, actions, and scenes, which contain more information compared to images. Subtitles are often in spoken language, which contains inherent ambiguities due to multiple potential interpretations~\cite{DBLP:journals/corr/abs-2211-12503}. Hence, our dataset can be a complement to existing MMT datasets.

\section{Dataset}
\label{dataset}

We present \method, consisting of 150 thousand unique videos (9,981 hours in total) with both English and Chinese subtitles. The videos are collected from two popular online video platforms, YouTube and Xigua. All subtitles are human-written. Table~\ref{tab:basic_stats} lists statistics of our dataset and existing video-guided translation datasets. Among existing datasets, our dataset is significantly larger, with more videos and parallel sentences. 


\begin{table}[t]
    \centering
    \fontsize{10}{11}\selectfont
    \setlength{\tabcolsep}{1.0mm}
    \begin{tabular}{lrrrrrr}
    \toprule
    \textbf{Dataset} & \multicolumn{2}{c}{\textbf{Text}} & \multicolumn{3}{c}{\textbf{Video}} \\
    & \# Sent & Len. & \# Video & \# Clip & Sec. \\
    \midrule
    \textsc{VISA} & 35K & 7.0 & 2K & 35K & 10.0 \\
    \vatex & 129K & 15.2 & 25K & 25K & 10.0  \\
    \howtwo  & 186K & 20.6 & 13K & 186K & 5.8  \\
    \method & 4.5M & 22.8 & 156K & 4.5M & 8.0 \\
    \bottomrule
    \end{tabular}
    \caption{Statistics of the \method and existing video-guided machine translation datasets. The size of the \method bypasses the size of largest available datasets by one order of magnitude. For the text length (Len.), we report average length of source sentences (English) for fair comparison. For the video length (Sec.), we report average seconds of videos for fair comparison. }
    \label{tab:basic_stats}
\end{table}


\subsection{\method Dataset}

To obtain high-quality video-subtitle pairs, we collect videos with both English and Chinese subtitles from \textit{YouTube}\footnote{\href{https://www.youtube.com}{https://www.youtube.com}} and \textit{Xigua}\footnote{\href{https://www.ixigua.com}{https://www.ixigua.com}}. Both two platforms provide three types of subtitles: 1) \textit{creator} which is uploaded by the creator, 2) \textit{auto-generate} which is generated by the automatic speech recognition model and 3) \textit{auto-translate} which is produced by machine translation model. We only consider videos with both English and Chinese subtitles uploaded by creators in order to obtain high-quality parallel subtitles. These videos and subtitles are often created by native or fluent English and Chinese speakers. In total, we collect 13K videos (6K hours in total) from YouTube and 2K videos from Xigua (3.9K hours in total).

\noindent \textbf{Preprocessing.} We first re-segment English subtitles into full sentences. To ensure the quality of parallel subtitles, we use quality estimation scores (e.g., the COMET score) to filter out low-quality pairs. More details are provided in Appendix~\ref{sec:preprocessing_appendix}. Ultimately, 3.3M sentences paired with video clips are kept for YouTube, and 1.2M for Xigua. The average lengths of English and Chinese sentences are 17.6 and 15.4 words for YouTube, 37.7 and 32.4 words for Xigua. 


\begin{table}[t]
    \centering
    \fontsize{10}{11}\selectfont
    \setlength{\tabcolsep}{1.0mm}
    \begin{tabular}{lcc}
    \toprule
    \multirow{2}{*}{\textbf{Source}} & \multirow{2}{*}{\textbf{Fluency}$\uparrow$} & \textbf{Translation} \\
     &  & \textbf{Quality}$\uparrow$ \\
    
    \midrule
    \textsc{YouTube} & 4.81 & 4.11   \\
    \textsc{Xigua} & 4.60 & 4.20 \\
    \bottomrule
    \end{tabular}
    \caption{Human evaluation results on \method. The \textbf{fluency} score (1-5) measures whether source sentences (English) are fluent and error-free and the \textbf{translation quality} (1-5) measures whether sentence pairs are equivalent in meaning. Inter-annotator agreement with Krippendorf's $\alpha$ for all columns: 0.76, 0.82.}
    \label{tab:eval_dataset}
\end{table}


\subsection{Dataset Analysis}
\label{dataset_analysis}

\noindent \textbf{Quality Evaluation.} To assess the quality of text pairs, we randomly select 200 videos from each source and recruit seven annotators to rate the quality of subtitles pairs. For each video, we randomly select at most 20 clips for evaluation.
All annotators are fluent in English and Chinese. After watching video clips and subtitles, human annotators are asked to rate subtitle pairs from 1 (worst) to 5 (best) on \textbf{fluency}--whether the source sentence (English) is fluent and grammatically correct, and \textbf{translation quality}--whether the Chinese subtitle is semantically equivalent to the English subtitle. Detailed guidelines are provided in Appendix~\ref{sec:Annotation Guidelines}.

From Table~\ref{tab:eval_dataset},  English sentences from both YouTube and Xigua have an average of 4.8 and 4.6 fluency scores, which shows that English subtitles are fluent and rarely have errors. In terms of translation quality, we find more than 96 percent of the pairs are equivalent or mostly-equivalent, with only minor differences (e.g., style).

\begin{figure}[t]
    \centering
    \includegraphics[width=\columnwidth,trim=0 0 0 0, clip]{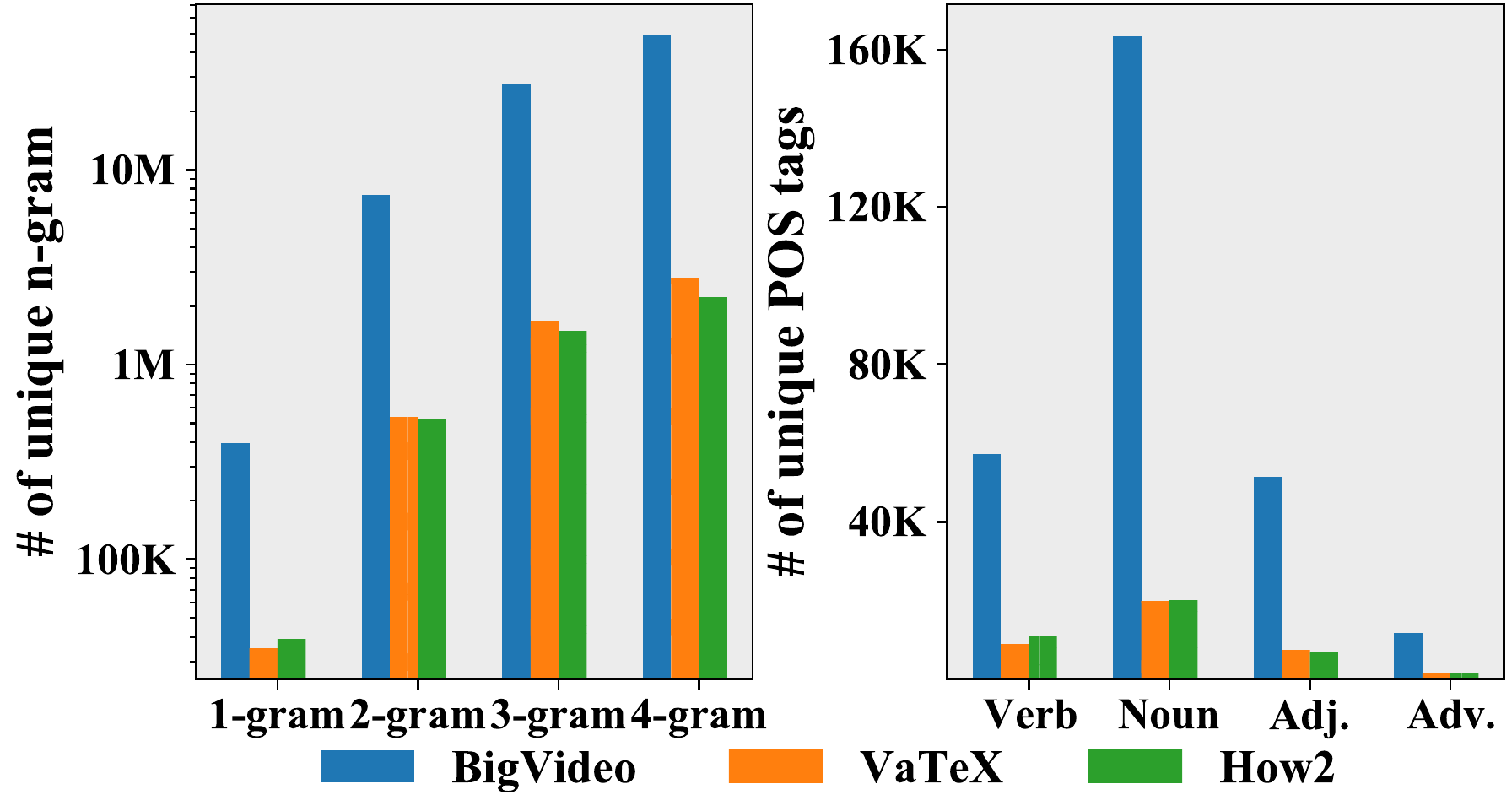}
    \caption{The numbers of unique $n$-grams and POS tags in \method exceed existing VMT datasets by one order. Our dataset is lexically richer than \vatex and \howtwo. For all datasets, we report statistics of source sentences (English) for fair comparison.}
    \label{fig:uniquengram}
\end{figure}

\noindent \textbf{Diversity Evaluation.} In addition to the size and quality, diversity is also critical for modeling alignments between parallel texts~\cite{tiedemann-2012-parallel}. Prior work calculates unique $n$-grams and part-of-speech (POS) tags to evaluate linguistic complexity~\cite{DBLP:conf/iccv/WangWCLWW19}. Besides word-level metrics, we use video category distribution to assess video-level diversity.

Since the source text of our dataset, \vatex and \howtwo are in English, we compare unique $n$-grams and POS tags on the source texts. For unique POS tags, we compare four most common types: verb, noun, adjective and adverb. In Figure~\ref{fig:uniquengram}, our data from both \textsc{Xigua} and \textsc{YouTube} have substantially more unique $n$-grams and POS tags than \vatex and \howtwo. Evidently, our dataset covers a wider range of actions, objects and visual scenes.

To evaluate video-level diversity, we compare category distributions among three datasets. The YouTube platform classifies videos into 15 categories. Since videos collected from the Xigua platform do not have category labels, we train a classifier on the YouTube data to label them. Details of the classifier are in Appendix~\ref{sec:Video Category Classifier Details}. Figure~\ref{fig:category_distribution} depicts the distributions of three datasets. While both \vatex and \howtwo have a long-tail distribution on several categories (e.g., ``Nonprofits \& Activism'' and ``News \& Politic''), \method has at least 1,000 videos in each category, which forms a more diverse training set.



\subsection{Test Set Annotation Procedure}
\label{testset_annotation}

\begin{table}[t]
    \centering
    \fontsize{10}{11}\selectfont
    \setlength{\tabcolsep}{2.0mm}
    \begin{tabular}{lccc}
    \toprule
    \textbf{Test set} & \textbf{Number} & \textbf{Length} & \textbf{\# phrases} \\
    \midrule
    \textsc{Ambiguous} & 877 & 28.61 & 745    \\
    \textsc{Unambiguous} & 1,517 & 27.22 & —  \\
    \bottomrule
    \end{tabular}
    \caption{Statistics of our test sets. We report number of samples, average length and number of ambiguous terms in two test sets. }
    \label{tab:test_statistics}
\end{table}

\begin{figure}[t]
    \centering
    \includegraphics[width=\columnwidth,trim=0 0 0 0, clip]{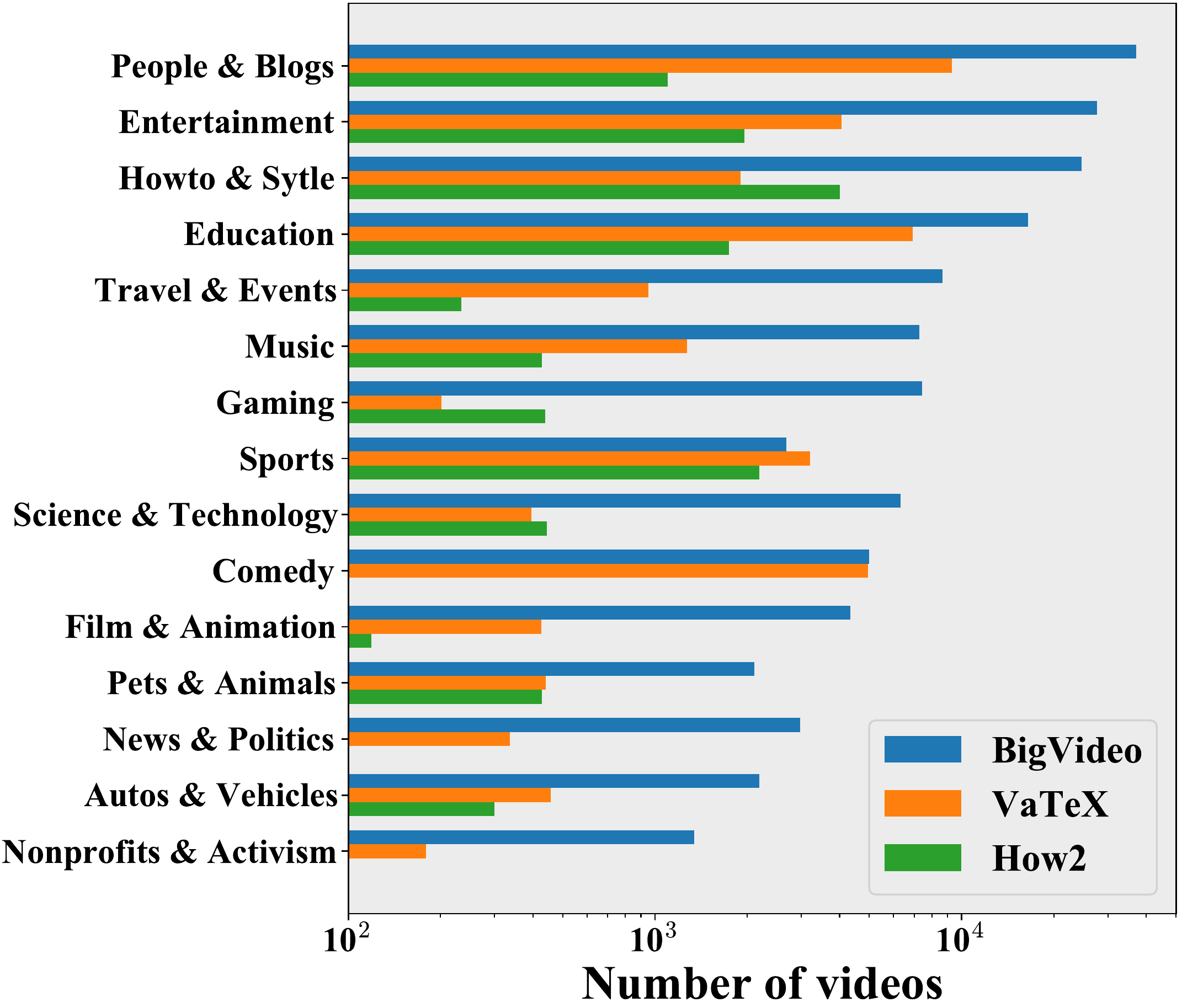}
    \caption{Category distribution on \method, \vatex and \howtwo. \method covers a wide range of domains.}
    \label{fig:category_distribution}
    \vspace{-2mm}
\end{figure}

Subtitles often contain semantic ambiguities ~\cite{gu-etal-2021-video}, which can be potentially solved by watching videos. In order to study ``\textit{How visual contexts benefit machine translation}'', we create two test sets: \testone contains 
ambiguous subtitles that videos provide strong disambiguation signal, while \testtwo consists of self-contained subtitles that videos are related but subtitles themselves contain enough context for translation. Statistics of two test sets are listed in Figure~\ref{fig:category_distribution}.

\begin{figure}[t]
    \centering
    \includegraphics[width=\columnwidth,trim=0 0 0 0, clip]{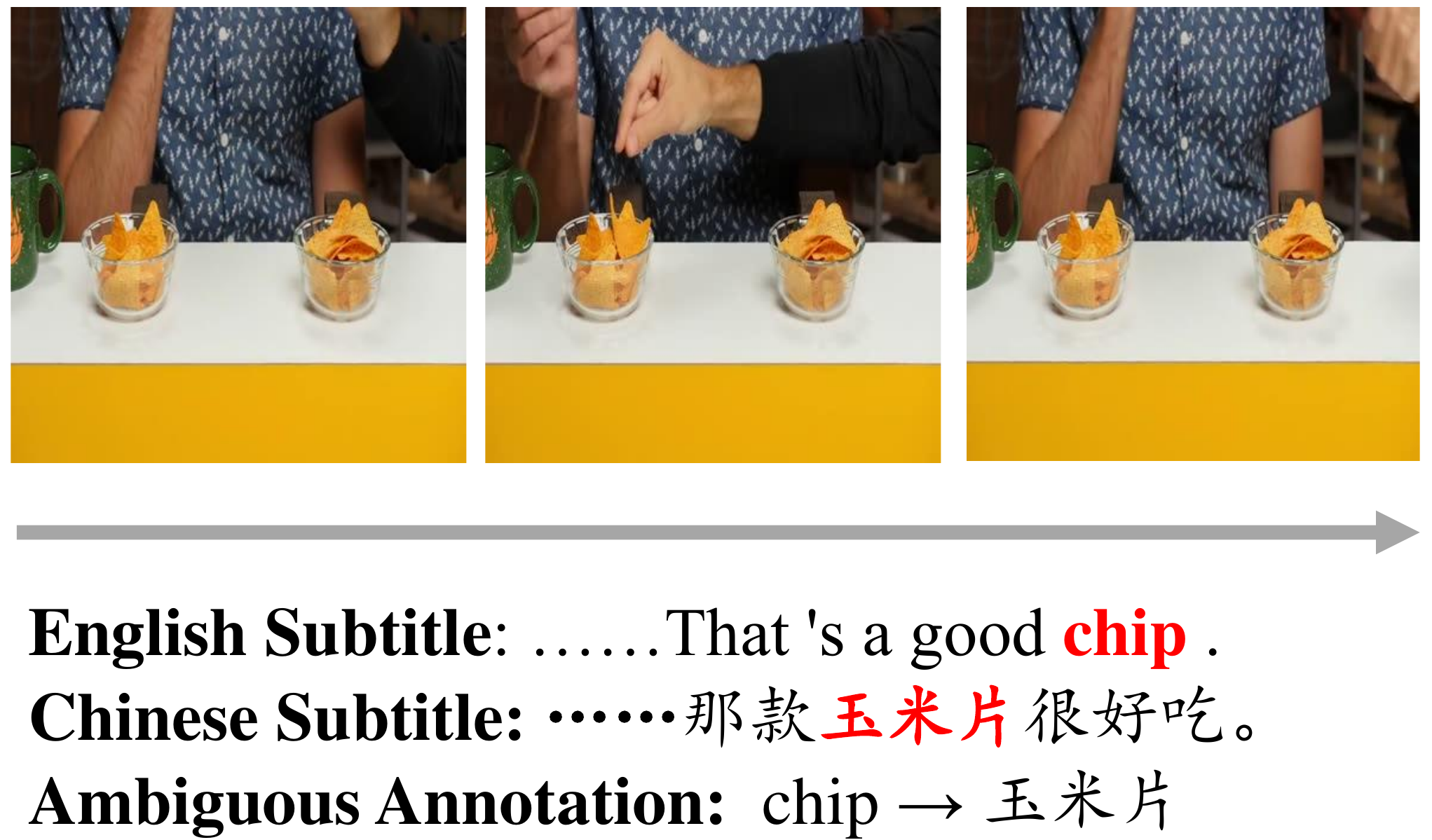}
    \caption{An example from our \testone test set. The ambiguous term ``chip'' is in \textcolor{red}{\textbf{red}}. }
    \label{fig:testset_example}
    \vspace{-2mm}
\end{figure}

We randomly sample 200 videos from each of Xigua and YouTube and hire four professional speakers in both English and Chinese to annotate the test set. Annotators are first asked to remove sentences which are not related to videos. In this step, we filter out about twenty percent of sentences. Annotators are then asked to re-write the Chinese subtitle if it is not perfectly equivalent to the English subtitle. Next, we ask the annotators to distinguish whether the source sentence contains semantic ambiguity. Specifically, annotators are instructed to identify ambiguous words or phrases in both English and Chinese sentences, as illustrated in Figure~\ref{fig:testset_example}. We finally obtain 2394 samples in our test set. 36.6\% of the sentences are in the \testone and 63.4\% of the sentences are in the \testtwo. In the \testone, we annotate 745 ambiguous terms. The statistics indicate that videos play important roles in our dataset. Annotation instructions and detailed procedures are provided in Appendix~\ref{sec:Annotation Guidelines}.


\section{Method}
\subsection{Model}
\begin{figure}[t]
	\begin{center}
		\includegraphics[width=\columnwidth]{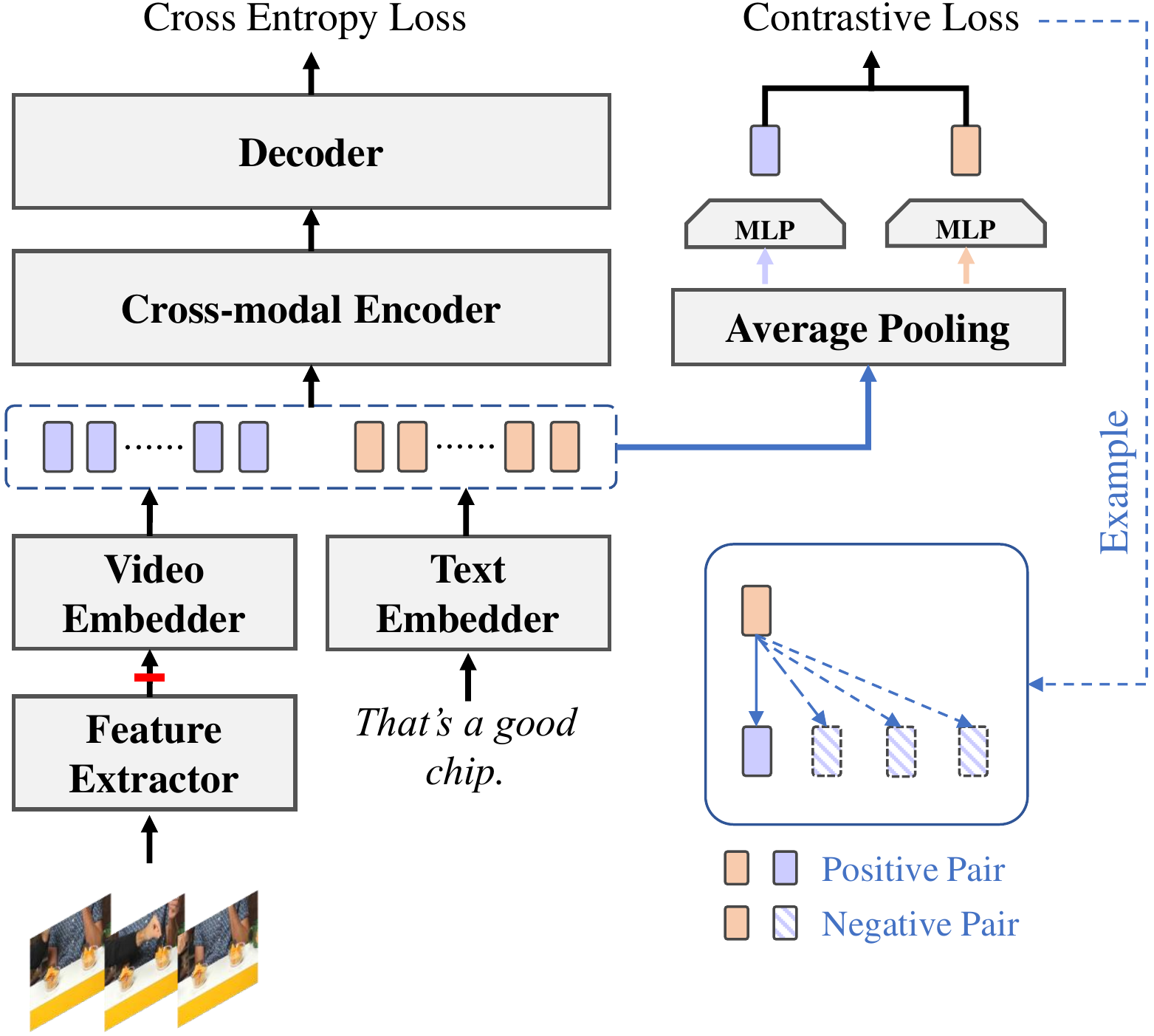}
	\end{center}
	\caption{An illustration of our machine translation system. An example of our contrastive learning is presented in the blue box.
	}
	\label{fig_modelframework}
\end{figure}

To better leverage videos to help translation, we present our video-guided machine translation model, as displayed in Figure~\ref{fig_modelframework}. Our model can be seamlessly plugged into the pretrained NMT model, which can benefit from large-scale parallel training data. Importantly, we design a contrastive learning objective to further drive the translation model to learn shared semantics between videos and text.

\noindent \textbf{Cross-modal Encoder.} Our model takes both videos and text as inputs. Text inputs are first represented as a sequence of tokens $\mathbf{x}$ and then converted to word embeddings through the embedding layer. Video inputs are represented as a sequence of continuous frames $\mathbf{v}$. We use a pretrained encoder to extract frame-level features, which is frozen for all experiments. Concretely, we apply the linear projection to obtain video features with the same dimension as text embeddings. To further model temporal information, we add positional embeddings to video features, followed by the layer normalization. Video features $\mathbf{v}^{emb}$ and text embeddings $\mathbf{x}^{emb}$ are then concatenated and fed into the Transformer encoder.

\noindent \textbf{Text Decoder.} Our decoder is the original Transformer decoder, which generates tokens autoregressively conditioned on the encoder outputs. We consider the cross entropy loss as a training objective:
\begin{equation}
    \begin{aligned}
	\mathcal{L}_{\rm CE} &= -\sum_{i}^{N} \log P(\mathbf{y}_{i}|\mathbf{v}_{i},\mathbf{x}_{i}), \\
    \end{aligned}
\end{equation} 
where $\mathbf{y}_i$ denotes the text sequence in the target language for the $i$-th sample in a batch of $N$ samples.

\subsection{Contrastive Learning Objective}
In order to learn shared semantics between videos and text, we introduce a cross-modal contrastive learning (CTR)-based objective. The idea of the CTR objective is to bring the representations of video-text pairs closer and push irrelevant ones further. 

Formally, given a positive text-video pair ($\mathbf{x}_i, \mathbf{v}_i$), 
we use remaining $N-1$ irrelevant text-video pairs ($\mathbf{x}_i, \mathbf{v}_j$) in the batch as negative samples. The contrastive learning objective ~\cite{DBLP:conf/nips/Sohn16} is:
\begin{equation}
    \begin{aligned}
	\mathcal{L}_{\rm CTR} = - \sum_{i=1}^{N} \log \frac{{\rm exp}(sim(\mathbf{x}_i^{p},\mathbf{v}_i^{p})/\tau)}{\displaystyle\sum_{j=1}^{N}{\rm exp}(sim(\mathbf{x}_i^{p},\mathbf{v}_j^{p})/\tau)}, \
    \end{aligned}
\end{equation} 
where $\mathbf{x}_i^{p}$ and $\mathbf{v}_i^{p}$ are representations for the text and video, $sim(\cdot)$ is the cosine similarity function and the temperature $\tau$ is used to control the strength of penalties on hard negative samples \cite{DBLP:conf/cvpr/WangL21a}.

\noindent \textbf{Text and Video Representations.} Importantly, since videos and subtitles are weakly aligned on the temporal dimension~\cite{DBLP:conf/iccv/MiechZATLS19}, we first average video embeddings and text embeddings in terms of the time dimension. Concretely, we apply two \textit{projection heads} ("MLP" in Figure~\ref{fig_modelframework}) to map representations to the same semantic space~\cite{DBLP:conf/icml/ChenK0H20}.

In the end, we sum up the two losses to obtain the final loss:
\begin{equation}
    \begin{aligned}
	\mathcal{L} &= \mathcal{L}_{\rm CE} + \alpha \mathcal{L}_{\rm CTR}, \\
    \end{aligned}
\end{equation} 
where $\alpha$ is a hyper-parameter to balance the two loss items.

\begin{table*}[!t]\small
	\renewcommand    
	\arraystretch{1.3}
	\centering
        \resizebox{\linewidth}{!}{
            \setlength{\tabcolsep}{1.5mm}
    	\begin{tabular}{l c c c c c c c c c}
    		\toprule
    		  \multirow{2}{*}{\textbf{System}} & \multicolumn{3}{c}{\textbf{BLEU}} & \multicolumn{3}{c}{\textbf{COMET}} & \multicolumn{3}{c}{\textbf{BLEURT}} \\
    		  \cmidrule(r){2-4}  \cmidrule(r){5-7}  \cmidrule(r){8-10}
    		 ~ & \textbf{All} & \textbf{Amb.} &  \textbf{Unamb.} & \textbf{All} & \textbf{Amb.} &  \textbf{Unamb.}  & \textbf{All} & \textbf{Amb.} &  \textbf{Unamb.}   \\
            \hline

          \multicolumn{4}{l}{\textit{w/o pretraining}} & & & & & &  \\
               \textsc{Text-only}  & $43.97$  & $43.59$   & $44.19$  & $37.31$  & $33.06$  & $39.77$  & $61.11$  & $59.68$  & $61.93$  \\
             \textsc{Gated Fushion}  & $44.33$  & $44.12$ & $44.45$ & $37.88$ & $34.47$ & $39.85$ & $61.25$ & $60.19$ & $61.86$ \\
             \textsc{Selective Attn}  & $44.39$ & $44.20$ & $44.51$ & $\mathbf{38.48}$ & $34.84$ & $\mathbf{40.59}$ & $\mathbf{61.37}$ & $60.30$ & $\mathbf{62.07}$  \\
            \textbf{Ours} \\
   
            $+$ \textsc{ViT}  & $44.26$   & $44.10$   & $44.37$  & $38.13$ & $34.75$  & $40.08$  & $61.24$ & $60.29$  & $61.78$  \\
            $+$ \textsc{SlowFast}  & $44.21$  & $44.12$   & $44.26$  & $37.81$  & $34.99$  & $39.44$  & $61.22$ & $60.28$  & $61.77$  \\
            $+$ \textsc{ViT} $+$ \textsc{CTR}  & $\mathbf{44.45}$  & $\mathbf{44.48}$   & $44.40$  & $38.15$  & $\mathbf{35.76}$  & $39.54$  & $61.36$  & $\mathbf{60.72}$ & $61.73$  \\
          $+$ \textsc{SlowFast} $+$ \textsc{CTR}  & $44.44$  & $44.20$   & $\mathbf{44.58}$  & $38.37$  & $35.18$  & $40.22$  & $61.31$  & $60.41$ & $61.82$  \\
           \hline
           \multicolumn{4}{l}{\textit{w/ pretraining}} & & & & & & \\
           \textsc{Text-only} & $44.45$   & $43.89$   & $44.79$  & $38.36$  & $33.40$  & \colorbox{pink}{$\mathbf{41.23}$}  & $61.41$  & $59.85$  & $62.31$ \\
           $+$ \textsc{ViT} $+$ \textsc{CTR} & \colorbox{pink}{$\mathbf{44.83}$}   & \colorbox{pink}{$\mathbf{44.62}$ }   & \colorbox{pink}{$\mathbf{44.96}$}  & \colorbox{pink}{$\mathbf{39.44}$}  & \colorbox{pink}{$\mathbf{36.42}$}  & $41.19$  & \colorbox{pink}{$\mathbf{61.76}$}  & \colorbox{pink}{$\mathbf{60.75}$}  & \colorbox{pink}{$\mathbf{62.34}$} \\
            $+$ \textsc{SlowFast} $+$ \textsc{CTR}  & $44.77$  & $44.43$   & $44.97$  & $39.26$  & $36.03$  & $41.12$  & $61.71$  & $60.52$ & $62.40$  \\
    	\bottomrule
    	\end{tabular}
     }
	\caption{
		\label{Table_Main_Results}
	sacreBLEU(\%), COMET(\%) and BLEURT(\%) scores on \method testset. We report results on \testone (\textbf{Amb.}), \testtwo (\textbf{Unamb.}) and whole test set (\textbf{All}). ``$+$ CTR'' denotes our cross-model framework with the contrastive learning objective. All results are mean values of five different random seeds. Complete results with standard deviations can be seen in Appendix~\ref{sec:complete_resullts}. The Best result in each group is in \textbf{bold}. The Best result in each column is in \colorbox{pink}{\textbf{red}}.}
\end{table*}

\section{Experiments}
\subsection{Experimental Setup}

\noindent \textbf{Implementation Details.} We evaluate our method on three video translation datasets: \vatex, \howtwo and our proposed dataset \method. More dataset details can be found in Appendex~\ref{appendix:c1_dataset}. 

Our code is based on the \textit{fairseq} toolkit \cite{ott2019fairseq}. The Transformer-base model follows \cite{DBLP:conf/nips/VaswaniSPUJGKP17}. Both encoder and decoder have 6 layers, 8 attention heads, hidden size = 512, and FFN size = 2048. We utilize post-layer normalization for all models. On \vatex, we follow the Transformer-small setting from \citet{wu-etal-2021-good} for better performance, 6 layers for encoder/decoder, hidden size = 512, FFN size = 1024 and attention heads = 4. 

All experiments are done on 8 NVIDIA V100 GPUS with mixed-precision training   \cite{DBLP:conf/iclr/0002MMKAB0VKGHD18}, where the batch assigned to each GPU contains 4,096 tokens. More training details can be found in Appendix~\ref{appendix:training details}. We stop the training if the performance on the validation set does not improve for ten consecutive epochs. The running time is about 64 GPU hours for our system. During the inference, the beam size and the length penalty are set to 4 and 1.0.  We apply byte pair encoding (BPE) with 32K merge operations to preprocess sentences of our dataset. During training and testing, we uniformly sample a maximum of 12  frames as the video input. The text length is limited to 256. For the contrastive learning loss, we set $\alpha$ to 1.0 and $\tau$ to 0.002. The choices of hyper-parameters are in Appendix~\ref{appendix:the_choice_of_hyperparemeters}.

For video features, we extract 2D features and 3D features to compare their effects. Concretely, we experiment with two pretrained models to extract the video feature: 
\begin{inparaenum}[\it a)]
    \item The vision transoformer (\textsc{ViT})~\cite{DBLP:conf/iclr/DosovitskiyB0WZ21} which extracts frame-level features.
    \item The SlowFast model (\textsc{SlowFast}) which extracts video-level features~\cite{DBLP:conf/iccv/Feichtenhofer0M19}.
\end{inparaenum}
For 2D features, we first extract images at a fixed frame rate (3 frames per second). Then we utilize pretrained Vision Transformer\footnote{The model architecture is vit\_base\_patch16\_224.} (ViT)  to extract 2D video features into 768-dimensional vectors. Here the representation of $[CLS]$ token is considered as the global information of one frame. For 3D features, we extract 2304-dimensional SlowFast\footnote{SLOWFAST\_8x8\_R50.} features  at 2/3 frames per second.

\noindent \textbf{Baselines and Comparisons.} For baselines, we consider the base version of the Transformer (\textsc{Text-only}), which only takes texts as inputs. For comparisons, since most recent MMT studies focus on image-guided machine translation, we implement two recent image-based MMT models: 
\begin{inparaenum}[\it a)]
    \item The gated fusion model (\textsc{Gated Fusion}) which fuses visual representations and text representations with a gate mechanism~\cite{wu-etal-2021-good}.
    \item The selective attention model (\textsc{Selective Attn}) which uses a single-head attention to connect text and image representations~\cite{li-etal-2022-vision}.
\end{inparaenum}
We extract image features using ViT and obtain the visual feature by averaging image features on the temporal dimension. The visual feature is then fused with the text representations which is the same as original \textsc{Gated Fusion} and \textsc{Selective Attn}. For \howtwo and \vatex, we additionally include the baseline models provided by the original paper.

\noindent \textbf{Evaluation Metrics.} We evaluate our results with the following three metrics: detokenized sacreBLEU\footnote{{\href{https://github.com/mjpost/sacrebleu}{https://github.com/mjpost/sacrebleu}. \\Signature: BLEU+case.mixed+numrefs.1+smooth.exp+\\tok.zh+version.1.5.1}~\cite{post-2018-call}}, COMET\footnote{\href{https://github.com/Unbabel/COMET}{https://github.com/Unbabel/COMET}. We use the default wmt20-comet-da.}~\cite{rei-etal-2020-comet} and BLEURT\footnote{\href{https://github.com/google-research/bleurt}{https://github.com/google-research/bleurt}. We use the BLEURT-20.}~\cite{sellam-etal-2020-bleurt}. In order to evaluate whether videos are leveraged to disambiguate, we further consider three terminology-targeted metrics~\cite{DBLP:journals/corr/abs-2106-11891}:
\begin{itemize}
\setlength{\itemsep}{3pt}
\setlength{\parsep}{0pt}
\setlength{\parskip}{0pt}
    \item Exact Match: the accuracy over the annotated ambiguous words. If the correct ambiguous words or phrases appear in the output, we count it as correct.
    \item Window Overlap: indicating whether the ambiguous terms are placed in the correct context. For each target ambiguous term, a window is set to contain its left and right words, ignoring stopwords. We calculate the percentage of words in the window that are correct. In practice, we set window sizes to 2 (Window Overlap-2) and 3 (Window Overlap-3).
    \item Terminology-biased Translation Edit Rate (1-TERm): modified translation edit rate~\cite{DBLP:conf/amta/SnoverDSMM06} in which words in ambiguous terms are set to 2 edit cost and others are 1.

\end{itemize}

\section{Results}

\subsection{Main Results}

\noindent \textbf{Videos Consistently Improve the NMT Model.} As displayed in Table~\ref{Table_Main_Results}, on \method, our models equipped with videos obtain higher automatic scores. This indicates the benefit of using videos as additional inputs. Notably, our model trained with the additional contrastive learning objective yields better scores compared to the variant trained only  with the cross entropy loss. This signifies that our contrastive learning objective can guide better acquisition of video inputs. Furthermore, we find image-based pretrained model \textsc{ViT} and video-based pretrained model \textsc{SlowFast} yield comparable results, indicating that two vision features perform equally well on \method.

\begin{table}[t]\small
	\renewcommand
	\arraystretch{1.5}
         \resizebox{\linewidth}{!}{
	\centering
        \fontsize{8}{8}\selectfont
        \setlength{\tabcolsep}{0.5mm}
    	\begin{tabular}{l  c c c c }
    		\hline
    		  \multirow{2}{*}{\textbf{System}} & {\textbf{Exact}} & {\textbf{Window}} & {\textbf{Window}} & \multirow{2}{*}{{\textbf{1-TERm}}} \\

                                                 & {\textbf{Match}} & {\textbf{Overlap-2}} & {\textbf{Overlap-3}} &  \\
                           \hline
              \multicolumn{4}{l}{\textit{w/o pre-training}} \\
          \textsc{Text-only}  & $23.03$    & $14.22$  & $14.28$ & $49.50$  \\
         \textsc{Gated Fushion}  & $23.68$ &  $14.60$  & $14.76$ & $49.68$  \\
         \textsc{Selective Attn}  & $23.66$ &  $14.95$  & $15.08$ & $49.77$  \\

         Ours \\

         $+$ \textsc{ViT}    & $24.27$   & $15.09$ & $15.27$ & $49.78$  \\
          $+$ \textsc{SlowFast}  & $24.05$   & $14.97$ & $15.13$ & $49.32$ \\

           $+$ \textsc{ViT} $+$ \textsc{CTR}  & \colorbox{pink}{$\mathbf{25.02}$}   & \colorbox{pink}{$\mathbf{15.56}$} & \colorbox{pink}{$\mathbf{15.77}$} & $\mathbf{49.91}$  \\
            $+$ \textsc{SlowFast} $+$ \textsc{CTR} & $24.08$   & $15.04$ & $15.12$ & $49.32$  \\
    	\hline
            \multicolumn{4}{l}{\textit{w/ pre-training}} \\
            \textsc{Text-only} & $22.71$   & $14.35$ & $14.37$ & $49.73$    \\
            $+$ \textsc{ViT} $+$ \textsc{CTR}  &  $\mathbf{24.30}$   & $\mathbf{15.17}$ & $\mathbf{15.39}$ & \colorbox{pink}{$\mathbf{50.28}$}    \\
            $+$ \textsc{SlowFast} $+$ \textsc{CTR}  &  $23.62$   & $14.76$ & $14.90$ & $50.04$    \\
           
            \bottomrule
    	\end{tabular}
     }
	\caption{
		\label{Table_terminology_based}
	Terminology-targeted evaluation on \testone test set. Complete results with standard deviations can be seen in Appendix~\ref{sec:complete_resullts}. 
	}
\end{table}

Noticeably, compared to the text-only baseline, our models trained with the CTR objective achieves larger gain on \testone  than that on \testtwo. This demonstrates that it is more difficult to correctly translate sentences of \testone, while taking videos as additional inputs can help the model generate better translations.

To better study the role of videos in translation, we introduce additional training data to build a stronger NMT baseline. We introduce the WMT19 Zh-En dataset with 20.4M parallel sentences for pretraining. We aim to answer: \textit{how will the model perform if more text data are included?}

As displayed in Table~\ref{Table_Main_Results}, \textit{Model with video inputs outperforms the strong NMT baseline}. Pretraining on large corpus benefits models on \method. However, we find improvements mainly come from the \testtwo. This shows that videos play more crucial roles in \testone, which suggests that \method can serve as a valuable benchmark for studying the role of videos in MMT research.

\noindent \textbf{Videos Help Disambiguation.} We further evaluate the model ability of disambiguation. We present results on terminology-targeted metrics in Table~\ref{Table_terminology_based}. First, our systems with video features achieve consistent improvements both on exact match and window overlap metrics compared to the text-only variant, indicating that models augmented by video inputs correctly translate more ambiguous words and place them in the proper contexts. It is also worth noticing that our system with pretraining achieves better scores compared to the strong text baseline, which further highlights the importance of video inputs. Moreover, we find it hard to correctly translate ambiguous words since the best exact match score is 25.02\%, which suggests that our \testone set is challenging.

\begin{table}[t]
    \centering
    \fontsize{10}{11}\selectfont
     \resizebox{\linewidth}{!}{
    \begin{tabular}{l cccc}
    \toprule
    \textbf{Systems} & \textbf{Score} & \textbf{Win} $\uparrow$ & \textbf{Tie} & \textbf{Lose} $\downarrow$ \\
    \midrule
      \multicolumn{5}{c}{\testone} \\
        \midrule
    \textsc{Text-only} & $3.48$  &  --- & --- &  ---  \\
    $+$ \textsc{ViT} $+$ \textsc{CTR} & $\mathbf{3.53}$ & $19.3\%$ & $65.3\%$ & $15.3\%$ \\
         \midrule
     \multicolumn{5}{c}{\testtwo} \\
      \midrule
      \textsc{Text-only} & $3.71$  & --- & --- & ---    \\
      $+$ \textsc{ViT} $+$ \textsc{CTR} & $\mathbf{3.72}$ &  $24\%$ & $51.3\%$ & $21.7\%$ \\
    \bottomrule
    \end{tabular}}
    \caption{Human evaluation results on randomly sampled set.  ``\textbf{Win}''/``\textbf{Tie}''/``\textbf{Lose}'' stands for the percentage of translations where our system is better than, tied with, or worse than the text-only system. Inter-annotator agreement with Krippendorf's $\alpha$ for the translation quality score is 0.53.}
    \label{Table_human_evaluation}
\end{table}

\paragraph{Video-augmented Model Improves Translation Quality.} We further conduct human evaluation to analyze the translation quality. We randomly pick 100 sentences from the \testone and the \testtwo respectively and recruit three human judges for evaluation. For each sentence, the judges read the source sentence and two candidate translations, which are from \textsc{Text-only} and our model $+$ \textsc{ViT} $+$ \textsc{CTR}. The judges are required to rate each candidate on a scale of 1 to 5 and pick the better one. Detailed guidelines are in Appendix~\ref{sec:Human Evaluation Guidelines}.

From Table~\ref{Table_human_evaluation}, we can see \textit{our system with video inputs are more frequently rated as better translation than the text-only model on both \testone and \testtwo test sets}. This echoes automatic evaluations and implies that taking videos as inputs improve translation quality. Moreover, overall scores on \testtwo are better than those on \testone, which demonstrates that \testone is more challenging.


\begin{figure}[t]
    \centering
    \includegraphics[width=\columnwidth,trim=0 0 0 0, clip]{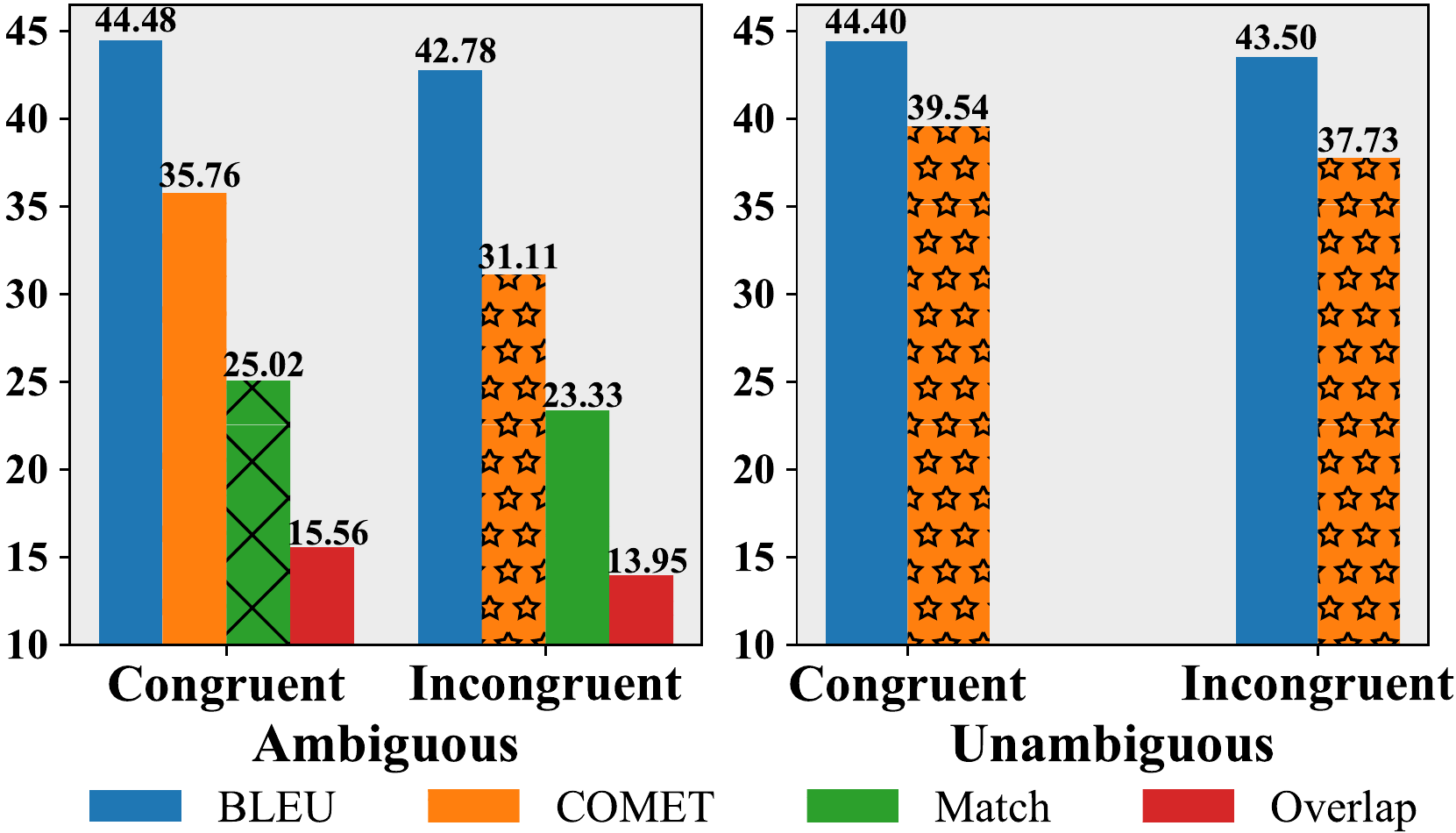}
    \caption{The results of incongruent decoding of the $+$ \textsc{ViT} $+$ \textsc{CTR} model. ``Congruent'' denotes the original results and ``Icongruent'' denotes the results of incongruent decoding where videos are replaced by wrong ones. ``Match'' represents the exact match score and ``Overlap'' stands for the ``Window Overlap-2''.
    \label{fig:incongruent-decoding} }
\end{figure}

\subsection{Incongruent Decoding}
In this section, we explore \textit{whether visual inputs contribute to the translation model}. Following~\cite{caglayan-etal-2019-probing,li-etal-2022-vision}, we use incongruent decoding to probe the need for visual modality on \method. During the inference, we replace the original video with a mismatched video for each sentence.  As shown in Figure \ref{fig:incongruent-decoding}, on \testone and \testtwo, we observe that all automatic metrics of our system drop significantly with incongruent decoding, suggesting the effectiveness of leveraging videos as inputs. 
Interestingly, we also find that the drop of the BLEU and COMET scores is larger on \testone than that on \testtwo, which further proves our point that videos are more crucial for disambiguation.

\subsection{Results on Public Datasets}

Next, we conduct experiments on public datasets, \vatex and \howtwo. Results are displayed in Table~\ref{Table_results_on_public}. On \howtwo, our best system achieves higher BLEU score 
 compared to the text-only model. However, the text-only model achieves best COMET and BLEURT, compared to all systems that take videos as inputs. On \vatex, our model with \textsc{SlowFast} features also achieves the highest scores on three evaluation metrics, compared to the text-only model and comparisons.  Notably, the model with \textsc{SlowFast} features is significantly better than models with \textsc{ViT} features, which is probably because \vatex focuses on human actions and the \textsc{SlowFast} model is trained on the action recognition dataset. However, the performance gap between the \textsc{Text-only} and our model $+$ \textsc{SlowFast} $+$ \textsc{CTR} is marginal. After we introduce 20M external MT data, we observe that the \textsc{Text-only} and our best system are comparable on automatic metrics. Since the cross-modal encoder often requires large-scale paired videos and text to train robust representations, our model does not achieve huge performance gain on \vatex and \howtwo. We hope our \method dataset can serve as a complement to existing video-guided machine translation datasets.

\begin{table}[t]\small
	\renewcommand
	\arraystretch{1.5}
	\centering
  \resizebox{\linewidth}{!}{
    	\begin{tabular}{l  c  c c c }
    		\hline
    		  {\textbf{System}} & {\textbf{BLEU}} & {\textbf{COMET}} & {\textbf{BLEURT}} \\
             \hline
             \multicolumn{4}{c}{\textsc{How2}} \\
             \hline
             \multicolumn{4}{l}{\textit{w/o pretraining}} \\
            \textsc{Text-only}  &  $57.57$  & \colorbox{pink}{$\mathbf{65.95}$}  & \colorbox{pink}{$\mathbf{72.52}$}  \\
    \citet{DBLP:journals/corr/abs-1811-00347} & $54.40$ & --- & --- \\
            \textsc{Gated Fushion}  &  $57.65$  & $65.12$  & $72.26$  \\
             \textsc{Selective Attn}  &  $57.51$  & $65.77$  & $72.35$  \\
         
                Ours \\

            $+$ \textsc{ViT} $+$ \textsc{CTR}  &  \colorbox{pink}{$\mathbf{57.95}$}  & $65.53$  & $72.46$    \\
              $+$ \textsc{SlowFast} $+$ \textsc{CTR}  &  $57.78$  & $65.58$  & $72.41$    \\
             \hline
             \multicolumn{4}{c}{\textsc{VaTeX}} \\
             \hline
             \multicolumn{4}{l}{\textit{w/o pretraining}} \\
               \textsc{Text-only} &  $35.01$  & $15.32$  & $56.99$   \\
             \citet{DBLP:conf/iccv/WangWCLWW19} &  $30.11$  & $4.50$  & $53.85$   \\
           \textsc{Gated Fushion } &  $33.79$  & $13.55$  & $55.66$ \\
            \textsc{Selective Attn}  &  $34.25$  & $13.55$  & $56.80$ \\
             Ours \\

              $+$ \textsc{ViT} $+$ \textsc{CTR}  & $34.84$  & $12.44$  & $56.25$      \\
              $+$ \textsc{SlowFast} $+$ CTR & $\mathbf{35.15}$  & $\mathbf{15.65}$  & $\mathbf{57.06}$     \\
              \multicolumn{4}{l}{\textit{w/ pretraining}} \\
              \textsc{Text-only} &  \colorbox{pink}{$37.57$}  & \colorbox{pink}{$\mathbf{25.22}$}  & \colorbox{pink}{$\mathbf{59.33}$}  \\
              $+$ ViT $+$ CTR &  $37.34$  & $24.07$  & $58.87$      \\
              $+$ \textsc{SlowFast} $+$ \textsc{CTR} & $37.58$  & $25.05$  & $59.20$  \\

    		\hline
    	\end{tabular}}
	\caption{
		\label{Table_results_on_public}
	Experimental results on \howtwo and \vatex. Complete results with standard deviations can be seen in Appendix~\ref{sec:complete_resullts}.
	}
\end{table}

\section{Conclusion}
In this paper, we present \method——a large-scale video subtitle Translation dataset for multimodal machine translation. We collect 155 thousand videos accompanied by over 4.5 million bilingual subtitles. Specially, we annotate two test subsets: \testone where videos are required for disambiguation and \testtwo where text contents are 
self-contained for translation. We also propose a cross-modal encoder enhanced with a contrastive learning objective to build cross-modal interaction for machine translation. Experimental results prove that videos consistently improve the NMT model in terms of the translation evaluation metrics and terminology-targeted metrics. Moreover, human annotators prefer our system outputs, compared to the strong text-only baseline. We hope our \method dataset can facilitate the research of multi-modal machine translation.

\section*{Limitations}
\method is collected from two video platforms Xigua and YouTube. All videos are publicly available. However, some videos may contain user information (e.g., portraits) or other sensitive information. Similar to \vatex and \howtwo, we will release our test set annotation and the code to reproduce our dataset. For videos without copyright or sensitive issues, we will make them public but limit for research, and non-commercial use (We will require dataset users to apply for access). For videos with copyright or sensitive risks, we will provide ids, which can be used to download the video. This step will be done under the instruction of professional lawyers.

Though we show that our model with video inputs helps disambiguation, we find that our model could yield incorrect translation due to the lack of world knowledge. For example, model can not distinguish famous table tennis player Fan Zhengdong and give correct translation. We find this is due to video pretrained models are often trained on action dataset (e.g., Kinetics-600~\cite{DBLP:conf/eccv/LongYQTL020}) and hardly learn such world knowledge. In this work, we do not further study methods that leverage world knowledge.

\section*{Ethical Considerations}

\noindent \textbf{Collection of \method.} We comply with the terms of use and copyright policies of all data sources during collection from the YouTube and Xigua platform. User and other sensitive information is not collected to ensure the privacy of video creators. The data sources are publicly available videos and our preprocessing procedure does not involve privacy issues. For all annotation or human evaluation mentioned in the paper, we hire seven full-time professional translators in total and pay them with market wage. All of our annotators are graduates.

\noindent \textbf{Potential Risks of \method and our model.} While \method consists of high-quality parallel subtitles, we recognize that our data may still contain incorrect samples. Our model may as well generate degraded or even improper contents. As our dataset is based on YouTube or Xigua videos, models trained on our dataset might be biased towards US or Chinese user perspective, which could yield outputs that are harmful to certain populations.

\section*{Acknowledgements}
The project was supported by the National Key Research and Development Program of China(No. 2020AAA0108004),
National Natural Science Foundation of China (No. 62276219) and Natural Science Foundation of Fujian Province of China (No. 2020J06001).
We also thank the reviewers for their insightful comments.



\normalem
\bibliography{acl2023}
\bibliographystyle{acl_natbib}

\appendix

\begin{table*}[!t]\small
	\renewcommand    
	\arraystretch{1.3}
	\centering
        \resizebox{\linewidth}{!}{
    	\begin{tabular}{l c c c c c c c c c}
    		\toprule
    		  \multirow{2}{*}{\textbf{System}} & \multicolumn{3}{c}{\textbf{BLEU}} & \multicolumn{3}{c}{\textbf{COMET}} & \multicolumn{3}{c}{\textbf{BLEURT}} \\
    		  \cmidrule(r){2-4}  \cmidrule(r){5-7}  \cmidrule(r){8-10}
    		 ~ & \textsc{All} & \textsc{Ambiguous} &  \textsc{Unambiguous} & \textsc{All} & \textsc{Ambiguous} &  \textsc{Unambiguous}  & \textsc{All} & \textsc{Ambiguous} &  \textsc{Unambiguous}  \\
            \hline

          \multicolumn{4}{l}{\textit{w/o pre-training}} & & & & & &  \\
             \textsc{Text-only}  & $43.97_{0.10}$  & $43.59_{0.29}$   & $44.19_{0.20}$  & $37.31_{0.34}$  & $33.06_{0.24}$  & $39.77_{0.56}$  & $61.11_{0.09}$  & $59.68_{0.21}$  & $61.93_{0.13}$  \\
             \textsc{Gated Fushion} $+$ \textsc{ViT} & $44.33_{0.13}$  & $44.12_{0.33}$ & $44.45_{0.12}$ & $37.88_{0.32}$ & $34.47_{0.62}$ & $39.85_{0.35}$ & $61.25_{0.07}$ & $60.19_{0.08}$ & $61.86_{0.08}$ \\
             \textsc{Selective Attn} $+$\textsc{ViT} & $44.39_{0.13}$ & $44.20_{0.22}$ & $44.51_{0.10}$ & $\mathbf{38.48_{0.19}}$ & $34.84_{0.39}$ & $\mathbf{40.59_{0.32}}$ & $\mathbf{61.37_{0.10}}$ & $60.30_{0.24}$ & $\mathbf{62.07_{0.13}}$  \\
             Ours \\
            $+$ \textsc{ViT}  & $44.26_{0.20}$   & $44.10_{0.24}$   & $44.37_{0.21}$  & $38.13_{0.65}$ & $34.75_{0.79}$  & $40.08_{0.71}$  & $61.24_{0.17}$ & $60.29_{0.30}$  & $61.78_{0.20}$  \\
            $+$ \textsc{SlowFast}  & $44.21_{0.17}$  & $44.12_{0.27}$   & $44.26_{0.27}$  & $37.81_{0.38}$  & $34.99_{0.25}$  & $39.44_{0.53}$  & $61.22_{0.09}$ & $60.28_{0.08}$  & $61.77_{0.17}$  \\
            $+$ \textsc{ViT} $+$ \textsc{CTR } & $\mathbf{44.45_{0.13}}$  & $\mathbf{44.48_{0.16}}$   & $44.40_{0.11}$  & $38.15_{0.56}$  & $\mathbf{35.76_{0.42}}$  & $39.54_{0.75}$  & $61.36_{0.17}$  & $\mathbf{60.72_{0.21}}$ & $61.73_{0.16}$  \\
          $+$ \textsc{SlowFast} $+$ \textsc{CTR}  & $44.44_{0.12}$  & $44.20_{0.12}$   & $\mathbf{44.58_{0.17}}$  & $38.37_{0.41}$  & $35.18_{0.71}$  & $40.22_{0.53}$  & $61.31_{0.10}$  & $60.41_{0.09}$ & $61.82_{0.14}$  \\
           \hline
           \multicolumn{4}{l}{\textit{w/ pre-training}} & & & & & & \\
           \textsc{Text-only} & $44.45_{0.11}$   & $43.89_{0.19}$   & $44.79_{0.16}$  & $38.36_{0.33}$  & $33.40_{0.29}$  & $\mathbf{41.23_{0.54}}$  & $61.41_{0.13}$  & $59.85_{0.16}$  & $62.31_{0.19}$ \\
           $+$ \textsc{ViT} $+$\textsc{CTR} & $\mathbf{44.83_{0.09}}$   & $\mathbf{44.62_{0.10}}$   & $44.96_{0.11}$  & $\mathbf{39.44_{0.51}}$  & $\mathbf{36.42_{0.41}}$  & $41.19_{0.89}$  & $\mathbf{61.76_{0.08}}$  & $\mathbf{60.75_{0.12}}$  & $62.34_{0.17}$ \\
           $+$ \textsc{SlowFast} $+$ \textsc{CTR}  & $44.77_{0.31}$  & $44.43_{0.15}$   & $\mathbf{44.97_{0.45}}$  & $39.26_{0.31}$  & $36.03_{0.52}$  & $41.12_{0.76}$  & $61.71_{0.17}$  & $60.52_{0.11}$ & $\mathbf{62.40_{0.25}}$  \\

    	\bottomrule
    	\end{tabular}}
	\caption{
		\label{Table_Main_Results_Appendix}
	Complete sacreBLEU(\%), COMET(\%) and BLEURT(\%) scores on \method testset.  ``$+$ CTR'' denotes our cross-model framework with contrastive learning loss. ``\textsc{ALL}'' represents the results on the whole test set. All results are mean values of five different random seeds. The best result in each group is in \textbf{bold}.
	}
\end{table*}
\begin{table}[t]\small
	\renewcommand
	\arraystretch{1.3}
         \resizebox{\linewidth}{!}{
	\centering
    	\begin{tabular}{l  c c c c }
    		\hline
    		  \multirow{2}{*}{\textbf{System}} & {\textbf{Exact}} & {\textbf{Window}} & {\textbf{Window}} & \multirow{2}{*}{{\textbf{1-TERm}}} \\

                                                 & {\textbf{Match}} & {\textbf{Overlap-2}} & {\textbf{Overlap-3}} &  \\
                           \hline
              \multicolumn{4}{l}{\textit{w/o pre-training}} \\
    	\textsc{Text-only}  & $23.03_{0.67}$    & $14.22_{0.41}$  & $14.28_{0.42}$ & $49.50_{0.13}$  \\
         \textsc{Gated Fushion}  & $23.68_{0.52}$ &  $14.60_{0.56}$  & $14.76_{0.47}$ & $49.68_{0.19}$  \\
         \textsc{Selective Attn}  & $23.66_{0.47}$ &  $14.95_{0.76}$  & $15.08_{0.65}$ & $49.77_{0.22}$  \\
         Ours \\
         $+$ \textsc{ViT}    & $24.27_{0.37}$   & $15.09_{0.47}$ & $15.27_{0.50}$ & $49.78_{0.41}$  \\
          $+$ \textsc{SlowFast}  & $24.05_{0.58}$   & $14.97_{0.77}$ & $15.13_{0.85}$ & $49.32_{0.44}$ \\
           $+$ \textsc{ViT} $+$ \textsc{CTR }  & $\mathbf{25.02_{0.74}}$   & $\mathbf{15.56_{0.65}}$ & $\mathbf{15.77_{0.55}}$ & $\mathbf{49.91_{0.20}}$  \\
            $+$ \textsc{SlowFast} $+$\textsc{CTR} & $24.08_{0.31}$   & $15.04_{0.14}$ & $15.12_{0.17}$ & $49.32_{0.18}$  \\
    	\hline
            \multicolumn{4}{l}{\textit{w/ pre-training}} \\
            \textsc{Text-only} & $22.71_{0.72}$   & $14.35_{0.54}$ & $14.37_{0.61}$ & $49.73_{0.21}$    \\
            $+$ \textsc{ViT} $+$ \textsc{CTR}  &  $\mathbf{24.30_{0.67}}$   & $\mathbf{15.17_{0.58}}$ & $\mathbf{15.39_{0.52}}$ & $\mathbf{50.28_{0.39}}$    \\
            $+$ \textsc{SlowFast} $+$ \textsc{CTR}  &  $23.62_{0.58}$   & $14.76_{0.48}$ & $14.90_{0.36}$ & $50.04_{0.13}$    \\
           
            \bottomrule
    	\end{tabular}}
	\caption{
		\label{Table_terminology_based_appendix}
	Complete terminology-targeted results on \method test set. All results are mean values of five different random seeds with standard deviations as subscripts. The Best result in each group is in \textbf{bold}.
	}
\end{table}
\begin{table}[t]\small
	\renewcommand
	\arraystretch{1.3}
	\centering
  \resizebox{\linewidth}{!}{
    	\begin{tabular}{l  c  c c c }
    		\hline
    		  {\textbf{System}} & {\textbf{BLEU}} & {\textbf{COMET}} & {\textbf{BLEURT}} \\
             \hline
             \multicolumn{4}{c}{\howtwo} \\
             \hline
             \multicolumn{4}{l}{\textit{w/o pretraining}} \\
             	\textsc{Text-only}  &  $57.57_{0.26}$  & $\mathbf{65.95_{0.71}}$  & $\mathbf{72.52_{0.24}}$  \\
         \citet{DBLP:journals/corr/abs-1811-00347} & $54.40$ & --- & --- \\
            \textsc{Gated Fushion}  &  $57.65_{0.35}$  & $65.12_{0.43}$  & $72.26_{0.27}$  \\
             \textsc{Selective Attn}  &  $57.51_{0.20}$  & $65.77_{0.93}$  & $72.35_{0.23}$  \\
                Ours \\
    
            $+$ \textsc{ViT} $+$ \textsc{CTR}  &  $\mathbf{57.95_{0.24}}$  & $65.53_{0.68}$  & $72.46_{0.25}$    \\
              $+$ \textsc{SlowFast} $+$\textsc{CTR}  &  $57.78_{0.09}$  & $65.58_{0.71}$  & $72.41_{0.15}$    \\
             \hline
             \multicolumn{4}{c}{\vatex} \\
             \hline
             \multicolumn{4}{l}{\textit{w/o pretraining}} \\
               \textsc{Text-only} &  $35.01_{0.14}$  & $15.32_{0.45}$  & $56.99_{0.307}$   \\
                          \citet{DBLP:conf/iccv/WangWCLWW19} &  $30.11_{0.72}$  & $4.50_{0.81}$  & $53.85_{0.55}$   \\
               \textsc{Gated Fushion } &  $33.79_{0.14}$  & $13.55_{0.42}$  & $55.66_{0.11}$ \\
            \textsc{Selective Attn}  &  $34.25_{0.30}$  & $13.55_{0.10}$  & $56.80_{0.11}$ \\
             Ours \\

              $+$ \textsc{ViT} $+$ \textsc{CTR}  & $34.84_{0.25}$  & $12.44_{0.64}$  & $56.25_{0.25}$      \\
              $+$ \textsc{SlowFast} $+$ \textsc{CTR} & $\mathbf{35.15_{0.24}}$  & $\mathbf{15.65_{0.35}}$  & $\mathbf{57.06_{0.06}}$     \\
              \multicolumn{4}{l}{\textit{w/ pretraining}} \\
              \textsc{Text-only} &  $37.57_{0.35}$  & $25.22_{0.66}$  & $59.33_{0.09}$  \\
              $+$ ViT $+$ CTR &  $37.34_{0.14}$  & $25.07_{1.21}$  & $58.87_{0.33}$      \\
              $+$ \textsc{SlowFast} $+$ \textsc{CTR} & $37.58_{0.15}$  & $25.05_{0.58}$  & $59.20_{0.13}$  \\
     
    		\hline
    	\end{tabular}}
	\caption{
		\label{Table_results_on_public_appendix}
	Complete experimental results on \howtwo and \vatex. All results are mean values of five different random seeds with standard deviations as subscripts. The Best result in each group is in \textbf{bold}.
	}
\end{table}

\section{Complete Results}
\label{sec:complete_resullts}
The Complete results with standard deviations can be found in Table~\ref{Table_Main_Results_Appendix}, Table~\ref{Table_terminology_based_appendix} and Table~\ref{Table_results_on_public_appendix}.

\section{Data Collection}

\subsection{Preprocessing}
\label{sec:preprocessing_appendix}
Subtitles are organized as a list of text chunks. Each chunk contains both English and Chinese lines and a corresponding timestamp. To obtain complete sentences, we start processing subtitles by merging chunks. Since English subtitles are often with strong punctuation marks, we greedily merge continuous segments (The start time of the second segment and the end time of the first segment are within 0.5 seconds) until an end mark is met at the end of the segment. To preserve context, we keep merging continuous sentences until a maximum time limit of 15 seconds is reached. Finally, we pair each merged segment with the video clip from the time interval corresponding to the segment.

English sentences from both YouTube and Xigua have an average of 4.6 fluency score, which shows that English subtitles are fluent and rarely have errors. In terms of translation quality, subtitles collected from Xigua have an average of 4.2 translation quality score, which indicates most of the subtitle pairs are equivalent or near-equivalent. In YouTube data, we find about 20 percent of sentence pairs are not equivalent or have major errors such as mistranslation or omission.

To remove low-quality pairs, we try three commonly-used quality estimation scores: 1) the COMET score, 2) the Euclidean distance based on the multilingual sentence embedding~\cite{artetxe-schwenk-2019-massively}, and 3) the round-trip translation BLEU score~\cite{moon-etal-2020-revisiting}. We filter out pairs if more than one score is lower than the threshold (set to 0.1, 4 and 20, respectively). On annotated samples, the average translation quality reaches 4.1 after cleaning.

\subsection{Video Category Classifier Details}
\label{sec:Video Category Classifier Details}
To construct a large-scale video-guided dataset, we collect videos from a variety of domains and categorized them into 15 classes based on their video categories in YouTube. We use the official youtube-dl\footnote{\href{https://youtube-dl.org}{https://youtube-dl.org}} toolkit to retrieve video categories and other metadata from YouTube. To ensure consistency between YouTube and Xigua videos, we train a category classifier to classify the video category tags of Xigua videos and those YouTube videos whose category information are missing. We train the category classifier using the English subtitles and category information of pre-labeled videos and use it to predict the category tags for the rest of the unlabeled videos. Specifically, we first group consecutive subtitles in a video by 5 and then concatenate them as input for the classifier. During inference, we predicted the category tags of groups in each video and obtain the video's label by voting. The category classifier model was fine-tuned based on the pre-trained \texttt{XLNet-large-cased} model, which performed well on other classification tasks such as XNLI \cite{NEURIPS2019_dc6a7e65, conneau2018xnli}. A statistical summary of the video categories of the train set can be found in Figure~\ref{fig:category_distribution}.

In addition, we also count the category tags of the videos in the test set, as listed in Table \ref{tab:video_category_test}. Similar to the train set, we directly obtain the category tags for most YouTube videos directly from YouTube and predict the tags of the remaining videos using the category classifier mentioned earlier. The statistics for video categories show that the videos in our dataset are diverse in terms of domain, both in the training and test sets. These statistics for the video categories provide a more comprehensive view of \method.

\begin{table}[bt]
    \centering
    \fontsize{10}{11}\selectfont
    \setlength{\tabcolsep}{2.0mm}
    \begin{tabular}{lccc}
    \toprule
    \textbf{Category} & \textbf{Xigua} & \textbf{YouTube} & \textbf{All}  \\
    \midrule
    People \& Blogs          &   15   &   29  &   44  \\
    Entertainment            &   48   &   14  &   62  \\
    Howto \& Sytle           &   11   &   37  &   48  \\
    Education                &   33   &   3   &   36  \\
    Travel \& Events         &   14   &   4   &   18  \\
    Music                    &   0    &   0   &   0   \\
    Gaming                   &   13   &   6   &   19  \\
    Sports                   &   5    &   1   &   6   \\
    Science \& Tech          &   30   &   2   &   32  \\
    Comedy                   &   0    &   3   &   3   \\
    Film \& Animation        &   3    &   1   &   4   \\
    Pets \& Animals          &   12   &   6   &   18  \\
    News \& Politics         &   1    &   1   &   2   \\
    Autos \& Vehicles        &   9    &   2   &   11  \\
    Activism                 &   0    &   2   &   2   \\
    \midrule
    All                      &   194  &   111 &   305 \\
    \bottomrule
    \end{tabular}
    \caption{Video category tags of our test set. For most of the YouTube videos, we obtain the category tags from YouTube official. For the rest of the YouTube videos and all of the Xigua videos, we train a classifier and predict the category tags according to the subtitles.}
    \label{tab:video_category_test}
\end{table}

\begin{table}[bt]
    \centering
    \fontsize{10}{11}\selectfont
    \setlength{\tabcolsep}{2.0mm}
    \resizebox{\linewidth}{!}{
    \begin{tabular}{l c c c}
    \toprule
    \textbf{Hyperparameters} & \method & \howtwo & \vatex  \\
    \midrule
    GPUs           &  8  & 2 & 1  \\
    Batch Size     & 4,096 & 4,096 & 4,096 \\
    Dropout                &   0.1 & 0.3 & 0.3    \\
    Weight Decay         &   0.1   &   0.1   &   0.1    \\
    Learning Rate          &   7e-4   &   5e-4 &   1e-3   \\
    Warmup Steps             &   4000  &   4000 &   2000\\
    Layer Normalization      &  PostNorm  &  PostNorm &  PostNorm   \\
    \bottomrule
    \end{tabular} }
    \caption{Training hyperparameters details.}
    \label{Table_training_parameter}
\end{table}
\section{Experimental Detatils}

\subsection{Dataset}
\label{appendix:c1_dataset}
We additionally conduct experiments on two public video-guided translation datasets, \textsc{VaTeX} \cite{DBLP:conf/iccv/WangWCLWW19} and \textsc{How2} \cite{DBLP:journals/corr/abs-1811-00347}. The \textsc{How2} dataset is a collection of instructional videos from Youtube. The corpus contains 184,948 English-Portuguese pairs for training, each associated with a video clip. We utilize val (2,022) as the validation set and dev5 (2,305) as the testing set. The \textsc{VaTeX} dataset is a video-and-language dataset  containing over 41,250 unique videos. The released version of the bilingual collection includes 129,955 sentence pairs for training, 15,000 sentence pairs for validation, and 30,000 for testing. Since the testing set is not publicly available, we split the original validation set into two halves for validation and testing. Some video clips of \textsc{Vatex} are no longer available on the Youtube. So after removal, the used corpus contains 115,480 sentence pairs for training, 6,645 sentence pairs for validation, and 6,645 sentence pairs for testing, each associated with a video clip.

\begin{figure}[t]
    \centering
    \includegraphics[width=\columnwidth,trim=0 0 0 0, clip]{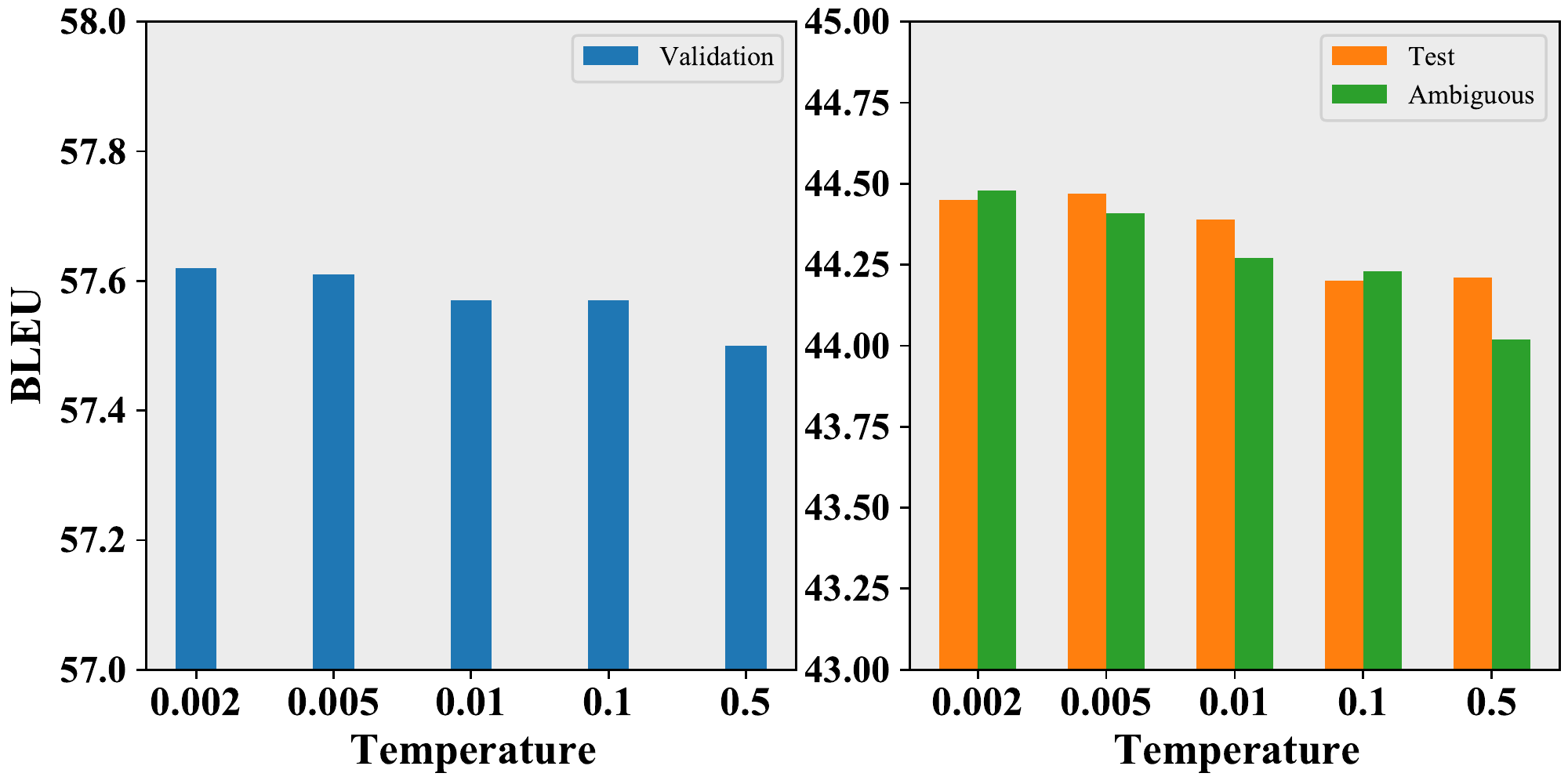}
    \caption{Bleu scores on \method validation, test and \testone sets. The x-axis is the choice of the different temperatures for the contrastive learning objective.
    \label{fig:temperature} }
\end{figure}

\begin{figure}[t]
    \centering
    \includegraphics[width=\columnwidth,trim=0 0 0 0, clip]{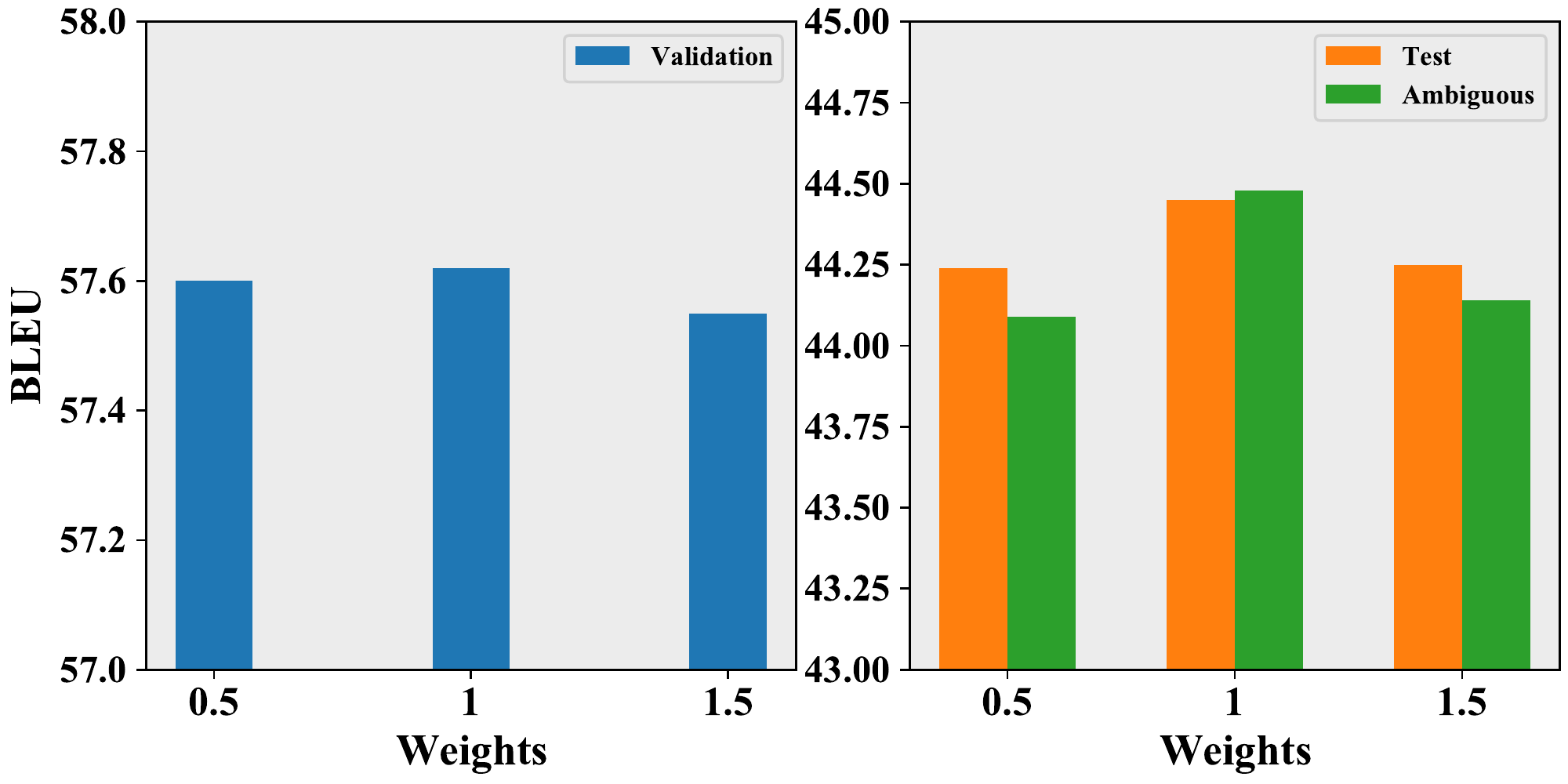}
    \caption{Bleu scores on \method validation, test and \testone sets. The x-axis is the choice of different weights for the contrastive learning objective.
    \label{fig:weight} }
\end{figure}

\begin{figure}[t]
    \centering
    \includegraphics[width=\columnwidth,trim=0 0 0 0, clip]{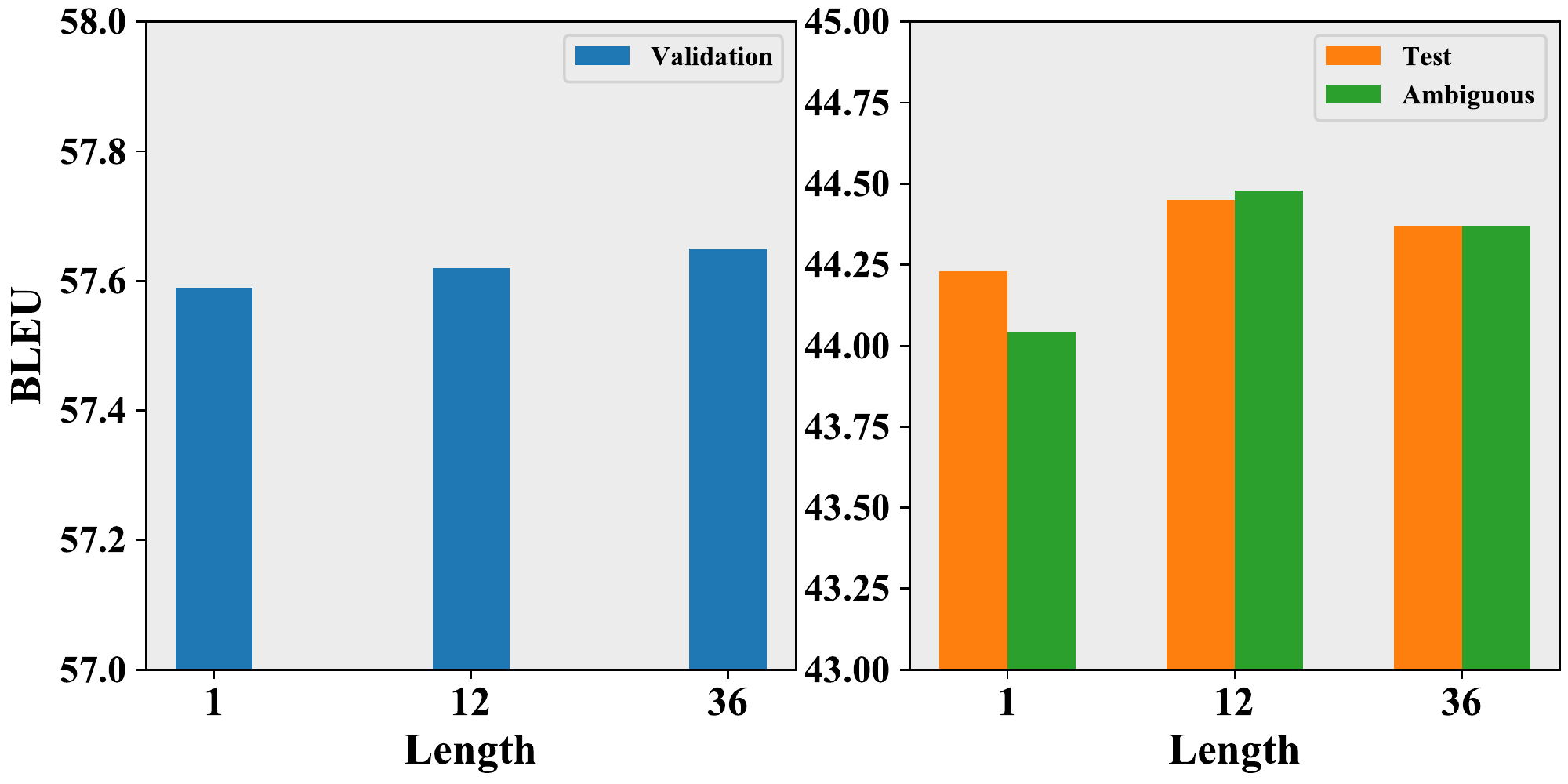}
    \caption{Bleu scores on \method validation, test and \testone sets. The x-axis is the length of video frames for our system.
    \label{fig:frame} }
\end{figure}

\subsection{Training and Implementation Details}
\label{appendix:training details}
More training details can be found in Table~\ref{Table_training_parameter}. For the pretraining on WMT19 Zh-En dataset, we utilize the same training parameters as that on \method and train the model for 300k steps.

\section{The Choice of Hyper-parameters}
\label{appendix:the_choice_of_hyperparemeters}

\noindent \textbf{Temperature for contrastive learning objective.} Performances of different temperature are presented in Figure~\ref{fig:temperature}. Here we fix the weight for contrastive learning objective to 1. On the validation set, there exists no significant difference in BLEU scores among choices of temperature. For better translation performance, a small temperature is more suitable.

\noindent \textbf{Weight for contrastive learning objective.} We fix  the $\tau=0.002$ and adjust the weight from 0.5 to 1.5. We can observe that contrastive learning objective with varying weights benefits the model to different degrees. 1.0 is the most suitable weight for our system.

\noindent \textbf{Length of Video Frames.} To investigate how the length of video frames affects translation, we adjust the number of sampled video frames in [1,12,36]. Figure~\ref{fig:frame} depicts their performances. Here the video features we use are 2D features extracted by ViT. We can observe that when only one video frame is sampled, the video degrades into one image and its positive impact on the system is reduced. A maximum of 12 video frames achieves the best performance.

\section{Case Study}
\begin{figure}[t]
    \centering
    \includegraphics[width=\columnwidth]{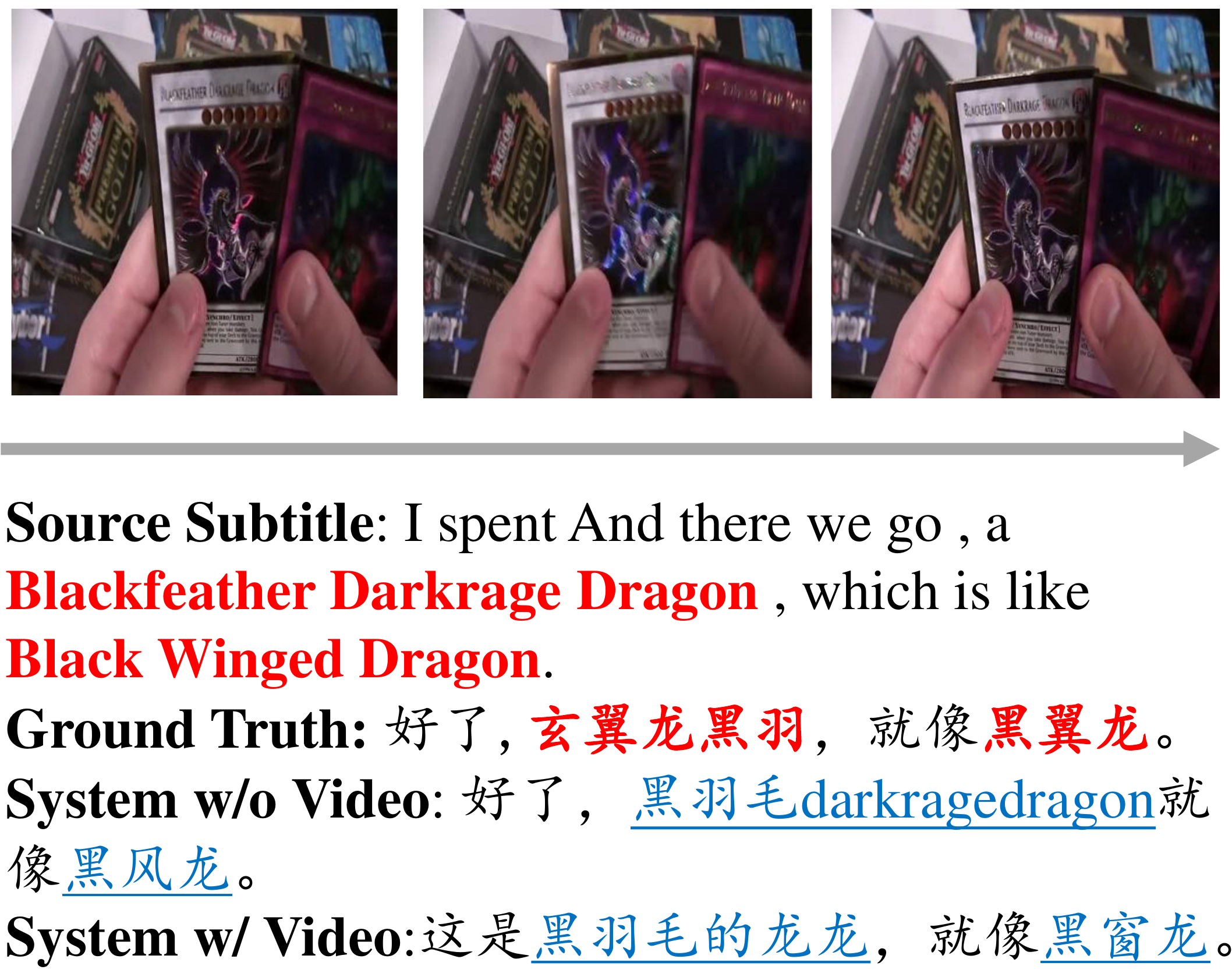}
    \caption{A case. The phrases with semantic ambiguity are highlighted in \textcolor{red}{\textbf{red}}. The wrong translations are in \textcolor[RGB]{0,112,192}{\underline{blue}} and the correct translations are in \textcolor[RGB]{191,144,0}{\textbf{yellow}}.}
    \label{fig:case_88}
\end{figure}

We additionally present two cases in the appendix. In figure~\ref{fig:case_522}, the phrase ``drive shot'' is better translated by our system by understanding the meaning of ``shot''. In Figure~\ref{fig:case_88}, we can find both the text-only baseline and our system fail to correctly translate the source title. The objects in the video are \textit{cards of Duel Monsters}, which need world knowledge to understand. So the source title is complicated for text-only and our system.

\begin{figure}[t]
    \centering
    \includegraphics[width=\columnwidth]{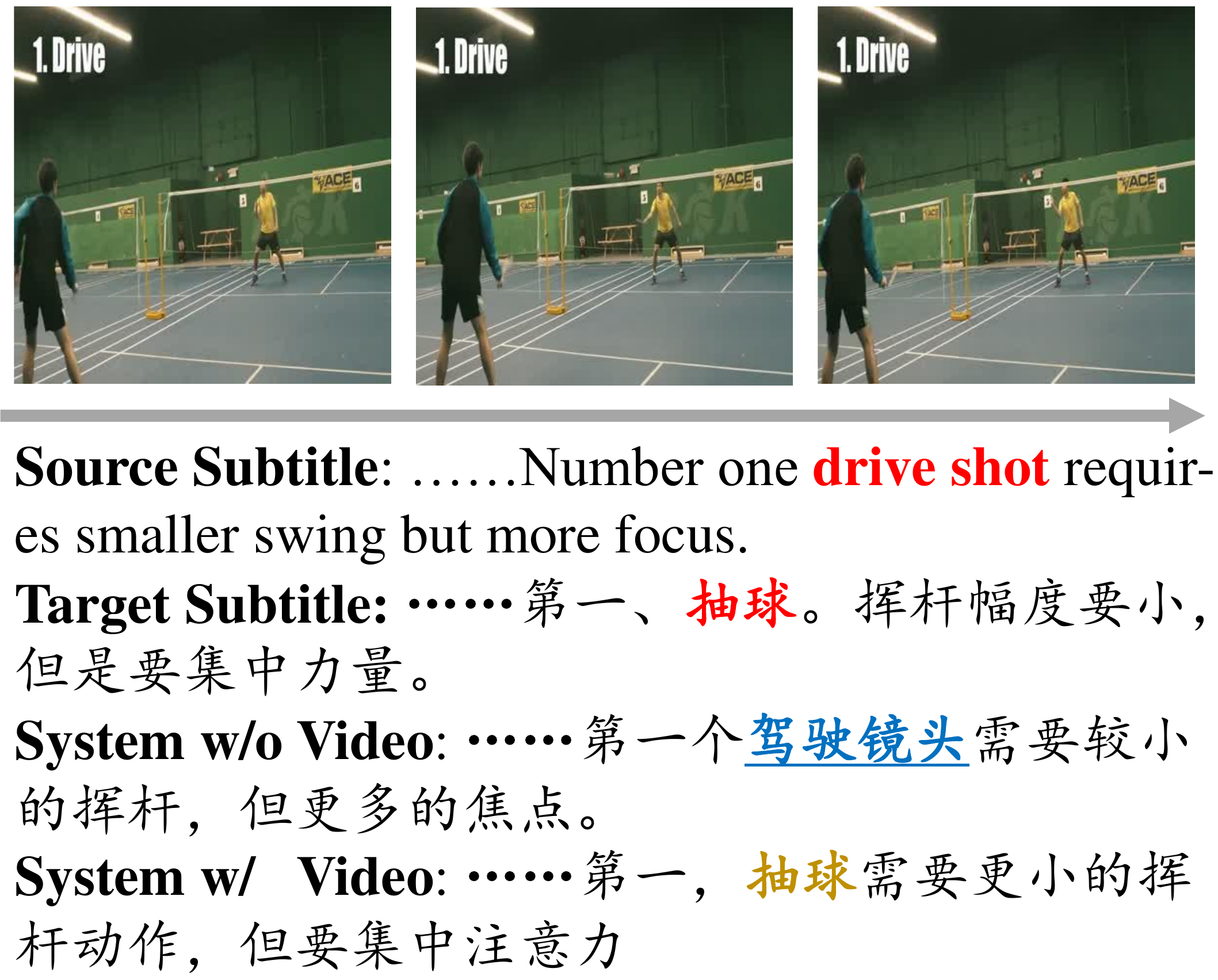}
    \caption{A case. The phrases with semantic ambiguity are highlighted in \textcolor{red}{\textbf{red}}. The wrong translations are in \textcolor[RGB]{0,112,192}{\underline{blue}} and the correct translations are in \textcolor[RGB]{191,144,0}{\textbf{yellow}}.}
    \label{fig:case_522}
\end{figure}

\section{Annotation Guidelines}
\label{sec:Annotation Guidelines}
We hire seven full-time annotators who are fluent in both Chinese and English. They are recruited to annotate translation data or conduct human evaluations. The annotators are shown one English and corresponding Chinese subtitle of the given video clip. After watching videos and reading subtitles, they are required to decide whether videos are related to subtitles. If not, the sample will be discarded. Then the annotators are required to rate on three aspects: 
\begin{itemize}
\setlength{\itemsep}{4pt}
\setlength{\parsep}{0pt}
\setlength{\parskip}{0pt}
\item \textbf{Fluency Score (1-5, 1 is the worst and 5 is the best):} If the audio is in English, the annotators will need to check whether the English subtitle is the transcript of the audio. If the audio is not in English, the annotators will need to rate if the sentence is grammatically correct.

\item \textbf{Translation Quality (1-5, 1 is the worst and 5 is the best):} Whether the Chinese subtitle is equivalent in meaning to the English Subtitle.

\item \textbf{Ambiguous (0/1):} The annotators need to decide whether the video information is required to complete the translation. "1" means "the video information is required" and otherwise "0".
\end{itemize}

\section{Human Evaluation Guidelines}
\label{sec:Human Evaluation Guidelines}
We hire three annotators to conduct the human evaluation. Each annotator is required to rate 100 samples from \testone and 100 samples from \testtwo on \textbf{translation quality} and rank two systems. The definition of the \textbf{translation quality} is the same as that in annotation guidelines.

\end{document}